\pdfoutput=1
\documentclass[journal]{IEEEtran}

\usepackage{times}
\usepackage{epsfig}
\usepackage{graphicx}
\usepackage{grffile}
\usepackage{amsmath}
\usepackage{amssymb}
\usepackage{placeins}

\usepackage{multirow}
\usepackage{multicol}
\usepackage{array}
\usepackage{rotating}
\usepackage[latin1]{inputenc}
\usepackage{color}

\usepackage{algorithmic}
\usepackage[boxed]{algorithm2e}
\algsetup{linenosize=\small}

\usepackage{grffile}
\usepackage{float}
\usepackage{changepage}

\ifCLASSOPTIONcompsoc
\else
\fi

\ifCLASSINFOpdf
\else
\fi

 \DeclareGraphicsExtensions{.eps,.ps,.pdf}
\graphicspath{{./figs/}}

\usepackage[breaklinks=true,bookmarks=false]{hyperref}

\begin{document}

\newcommand{\tab}[1]{Table~\ref{#1}}
\newcommand{\etal}{\emph{et al. }}
\newcommand{\ie}{i.e.} 
\newcommand{\eg}{e.g.} 
\newcommand{\fig}[1]{Fig.~\ref{#1}}
\newcommand{\subfig}[2]{Fig.~\ref{#1}#2}
\newcommand{\subtab}[2]{Table~\ref{#1}#2}
\newcommand{\points}{...,}
\newcommand{\sect}[1]{Sect.~\ref{#1}}
\newcommand{\eq}[1]{Eq. (\ref{#1})}
\newcommand{\iii}{{\cal I}}
\newcommand{\bfp}{{\bf p}}
\newcommand{\Algorithm}[1]{Algorithm~\ref{#1}}
\newcommand{\mys}[1]{{\footnotesize{#1}}}
\newcommand{\chap}[1]{Chap.~\ref{#1}}

\title{Road Detection by One-Class Color Classification: Dataset and Experiments}

\author{Jose~M.~\'Alvarez,
   Theo~Gevers,
   Antonio~M.~L\'opez
 \IEEEcompsocitemizethanks{\IEEEcompsocthanksitem
 Jose~M.~\'Alvarez is a researcher at NICTA, 2601 Canberra, Australia. jose.alvarez@nicta.com.au \IEEEcompsocthanksitem
 Antonio M. L\'opez is with the Computer Vision Center and the Computer Science
 Department, Univ. Aut\'onoma de Barcelona (UAB) 08193
 Bellaterra, Barcelona, Spain. \IEEEcompsocthanksitem Theo Gevers is
 with the Computer Vision Center (Barcelona) and the Intelligent System
 Laboratory Amsterdam, Faculty of Science, University of Amsterdam,
 Kruislaan 403, 1098 SJ, Amsterdam, The Netherlands.}
 \thanks{NICTA is funded by the Australian Government as represented by the Department of Broadband, Communications,
 and the Digital Economy, and the Australian Research Council (ARC) through the ICT Centre of Excellence Program.
 }
 }

\maketitle
\begin{abstract}
Detecting traversable road areas ahead a moving vehicle is a key
process for modern autonomous driving systems. A common approach
to road detection consists of exploiting color features to
classify pixels as road or background. These algorithms reduce the
effect of lighting variations and weather conditions by exploiting
the discriminant/invariant properties of different color
representations. Furthermore, the lack of labeled datasets has
motivated the development of algorithms performing on single
images based on the assumption that the bottom part of the image
belongs to the road surface.

In this paper, we first introduce a dataset of road images taken
at different times and in different scenarios using an onboard
camera. Then, we devise a simple online algorithm and conduct an
exhaustive evaluation of different classifiers and the effect of
using different color representation to characterize pixels.
\vspace{-0.25cm}
\end{abstract}

\section{Introduction}
\label{sect:Introduction} The main goal of vision-based road
detection is detecting traversable road areas ahead of an
ego-vehicle using an on-board camera. Detecting the course of the
road is a key component for the future development of driver
assistance systems and autonomous driving~\cite{Fritsch2014}. Road
detection using a monocular color camera is challenging since
algorithms must deal with continuously changing background, the
presence of different objects like vehicles and pedestrians,
different road types (urban, highways, country side) and varying
illumination and weather conditions. Moreover, these algorithms
should be executed in real-time.
\begin{figure}[t!]
\begin{center}
\includegraphics[width=\columnwidth]{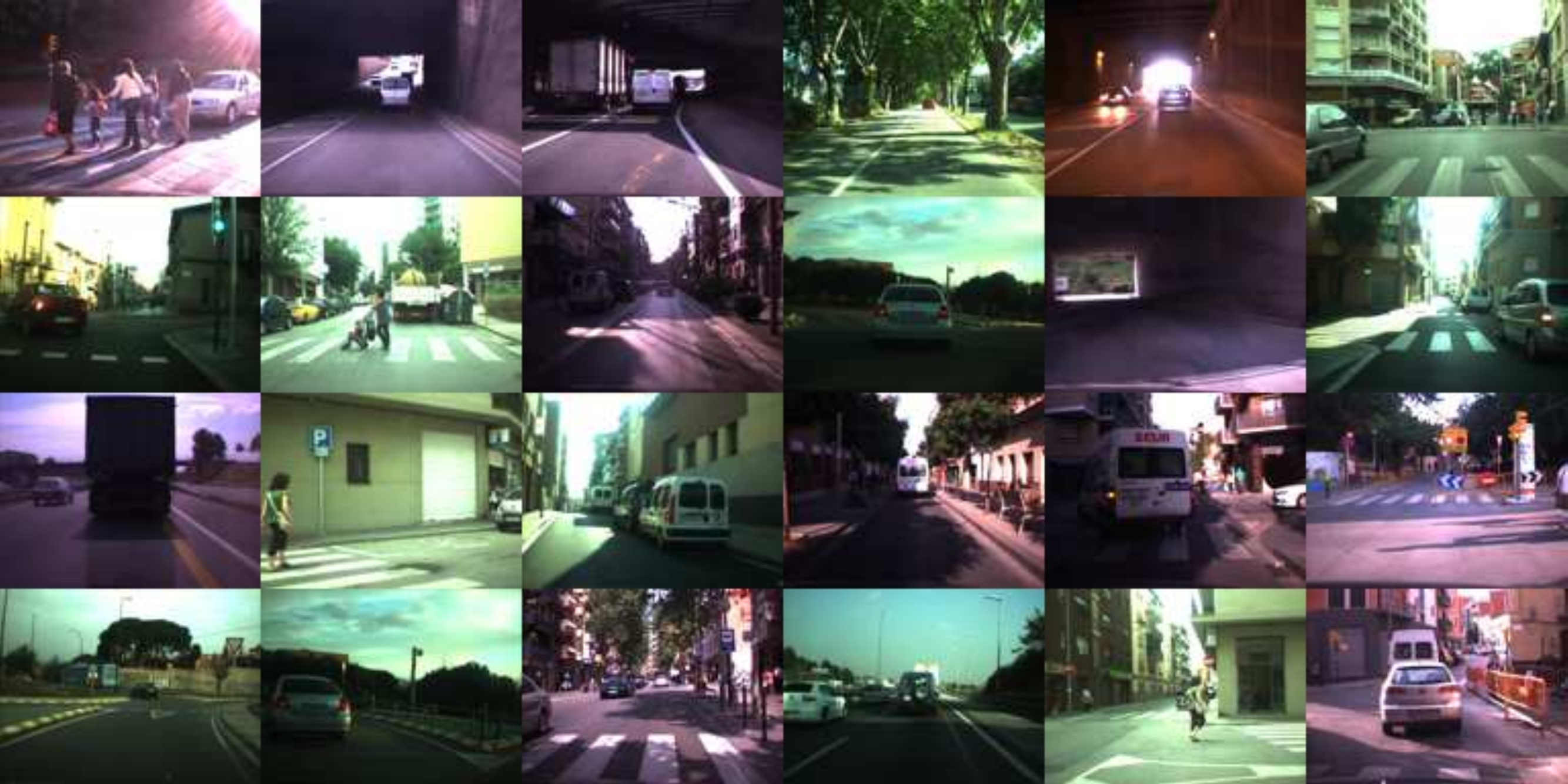}
\end{center}
\vspace{-0.3cm}
\caption{Example images of real driving situations. Online road
detection algorithms are especially suited for dealing with continuously changing conditions.}\label{fig:realdriving}
\vspace{-0.3cm}
\end{figure}

A common road detection approach consists of analyzing road
homogeneity to group pixels into road and background areas by
training a classifier based on road and non-road examples.
However, the large diversity of non-road areas and the lack of
annotated datasets hinder sampling these classes to create a
comprehensive representation~\cite{FritschITSC2014}. This has motivated the development
of algorithms that perform using only information from the current
image~\cite{SoteloITS:2004,Tan:2006,PonceTIP:2010,AlvarezITS:2011}.
These algorithms are usually referred as online road detection
algorithms. The core of these algorithms is a single class
classifier~\cite{TaxPhD:2001} trained on a small set of road
(positive) examples collected, for instance, from the bottom part
of the image being analyzed. Therefore, these algorithms do not
require samples of the background class. These algorithms are
highly adaptive to new road appearance and hence suited to deal
with constantly changing conditions that may occur in real driving
scenarios~(\fig{fig:realdriving}). In addition, to fit the real
time constraints, these algorithms usually represent pixel values
using simple (fast) features such as
color~\cite{SoteloITS:2004,Yinghua:2004,RodriguezIV2014,AlvarezIJCV:2010}
or texture~\cite{PonceTIP:2010,Lombardi:2005}. Color offers many
advantages over texture since texture varies with the speed of the
vehicle and with the distance to the camera due to perspective
effects. However, using color cues is a challenging task due to
the varying photometric conditions in the acquisition process
(\eg, illumination variations or different weather conditions).
Therefore, in this paper, we focus on evaluating online road
detection algorithms by using different single-class
classification methods performing on most common color
representations for online road detection. These different color
representations are evaluated on their robustness to varying
imaging conditions and their discriminative power. Moreover, a
road dataset is provided with ground truth to enable large scale
experiments for road detection. The dataset and the annotation
tool are made publicly available to the community at
http://scrd.josemalvarez.net/.

\begin{figure}[t!]
\begin{center}
\includegraphics[width=\columnwidth]{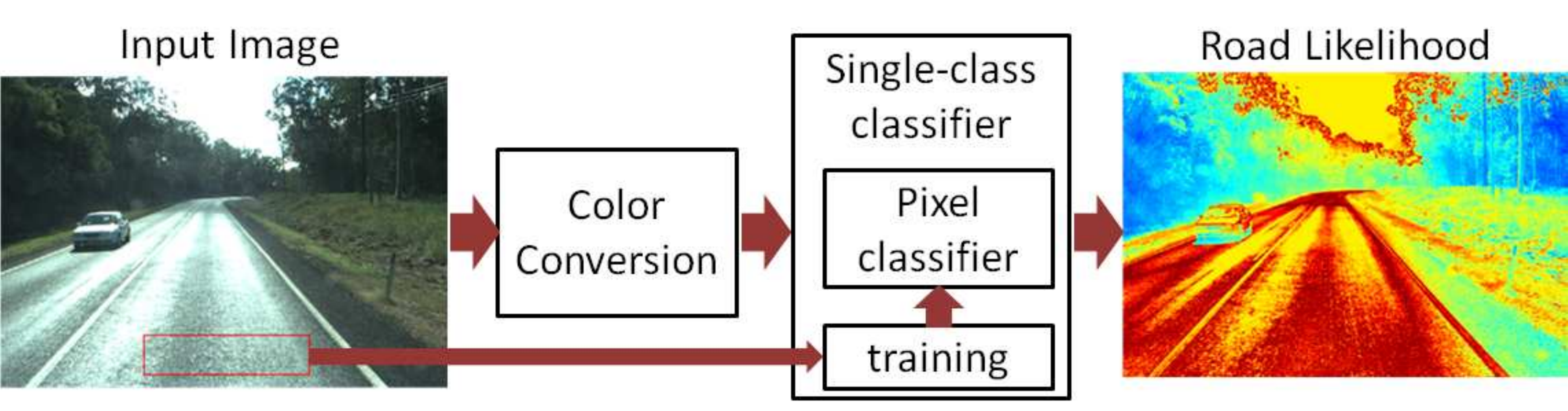}
\end{center}
\caption{Online road detection algorithm used in the experiments.
}\label{fig:algorithm}
\end{figure}

Hence, the contribution of this paper is two fold. First, we
provide a dataset to the community of on-board images for road
detection. The dataset consists of more than seven hundred
manually annotated images acquired at different daytime and
weather conditions in real world driving situations (urban
scenarios, highways and secondary roads). Second, we present a
comprehensive evaluation of existing single-class classifiers
using different color representations. To this end, we devise a
simple two stage algorithm for road detection for a single image.
In the first stage, the input image is converted to different
color representations and the output is used for a single class
classifier to provide a per-pixel confidence corresponding to the
probability of a pixel belonging to the road. The classifier is
trained using road pixels under the only assumption that a ROI in
the bottom part of the image belongs to the road surface,
see~\fig{fig:algorithm}.

The rest of this paper is organized as follows. First,
in~\sect{sect:sota} related work on road detection is reviewed.
Then, in~\sect{sect:algorithm}, we introduce the road detection
algorithms and the survey of color models and single
class-classifiers. The road dataset and the annotation tool are
introduced in~\sect{sect:BBDD}. Experiments are presented in
\sect{sect:experiments}. Finally, in \sect{sect:conclusions},
conclusions are drawn.

\section{Related Work}
\label{sect:sota} 
Common road detection methods analyze road
homogeneity by grouping pixels into road and background areas by
training a classifier based on road/non-road samples. However, the
large diversity of non-road areas and the lack of annotated
datasets has motivated the development of online detection
algorithms~\cite{SoteloITS:2004,Tan:2006,PonceTIP:2010,AlvarezWACV2014,AlvarezITS:2011,AlvarezIJCV:2010}.
The core of these algorithms is a single class classifier trained
on a small set of positive examples collected from the bottom part
of the image. Therefore, these algorithms do not require examples
of the background class. In addition, these algorithms represent
pixel values using simple (fast) features such as
color~\cite{SoteloITS:2004,Yinghua:2004} or
texture~\cite{PonceTIP:2010,AlvarezWACV2014} to be able to perform
in real-time. Color offers many advantages over texture since
texture varies with the distance to the camera. Color provides
powerful information about the road independent of the shape of
the road or perspective effects. However, using color cues is a
challenging task due to the varying photometric conditions in the
acquisition process. Different color planes exhibiting different
invariant properties have been used to reduce the influence of
these photometric variations. Color spaces derived from $RGB$ data
that have proved to be, to a certain extent, robust to lighting
variations are $HSV$~\cite{SoteloITS:2004,Rotaru:2008},
normalized $RGB$~\cite{Tan:2006}, CIE-$Lab$~\cite{Ess:2009} or
their combination~\cite{Ramstrom:2005,alvarezECCV:2012}. More
recently, color constancy has also been used
in~\cite{AlvarezITS:2011} to minimize the influence of lighting
variations. Algorithms embed these color representations in
complex systems that use inference methods (CRF), post-processing
steps and constraints such as temporal
coherence~\cite{Tan:2006,Michalke2009} or road shape
restrictions~\cite{SoteloITS:2004}. Therefore, is difficult to
compare and, more importantly, it is difficult to analyze
separately the different color representations to deal with
illumination changes within the road detection context.

\section{Online Road Detection Algorithm}
\label{sect:algorithm}

In this section, we present a simple framework for online road
detection. The algorithm, depicted in~\fig{fig:algorithm}, is
devised for still images and consists of two stages: color
conversion (\sect{sect:LowLevelCues}) and pixel classification
(\sect{subsect:classification}). In short, this algorithm performs
as follows: $RGB$ pixel values are converted to a preferred color
representation and then used as input to the classification stage.
The second stage is a single class classifier that considers only
road samples collected from the bottom part of the image. Thus,
the algorithm is based on the assumption that the bottom region of
the image belongs to the road class. This area usually corresponds
to a distance of about four meters ahead the camera and it is a
reasonably assumption when the car is on the road. The output of
the classifier is a road likelihood showing the probability of
each pixel of belonging to the road class defined as
$\mathcal{L_*}\in\mathbb{R}^N$ for a test image of $N$ pixels.
This likelihood ranges from $0$ to $1$ in which the higher the
likelihood, the higher the probability of being a road pixel.
State of the art algorithms build upon this road likelihood to
obtain the traversable road area incorporating post-processing
steps such as connected components~\cite{AlvarezITS:2011},
temporal coherence~\cite{Tan:2006,Michalke2009}, shape
restrictions~\cite{SoteloITS:2004} or even conditional random
fields results in robustified algorithms~\cite{ChunzhaoMRF:2010}.
In this paper, for fair comparison, we use a simple threshold to
assign pixel labels: if $\mathcal{L}_i > \tau$, the i-th pixel is
labeled by a road label. Otherwise, a background label is
assigned.

\subsection{Color Conversion}
\label{sect:LowLevelCues} The first stage is the color conversion
process to represent $RGB$ pixel values by different color models.
Algorithms have exploited several invariant/sensitive properties
of existing color spaces to reduce the influence of lighting
variations in outdoor scenes. In this paper, we analyze the
performance of five different device independent color spaces:
$RGB$ and the following four other
spaces~(\tab{tab:colorplanedefinition}): normalized $RGB$,
opponent color space $O_1O_2O_3$, $HSV$ and the CIE-$Lab$ space.
Each of these color spaces have different properties as summarized
in~\tab{tab:taxonomy}. For instance, color channels $R$, $G$, or
$B$ provide high discriminative power but limited invariance to
shadows and lighting variations. On the other hand, using hue $H$
or saturation $S$ to represent pixels provide higher invariance
but lower discriminative power. Three of the color spaces consider
separating the luminance and the chrominance into different
signals. For instance, in the $HSV$ color space, the $V$ channel
provides discriminative power while $H$ and $S$ components provide
different levels of invariance. Similarly, the opponent color
space comprises the luminance component and two channels providing
chromaticity information. As a result, these color representations
are uncorrelated and provide diversified color information.

\begin{table}[t!]
\caption{Derivation of the different color spaces used to characterize road pixels.}
\label{tab:colorplanedefinition}
{\small
\begin{minipage}[b]{0.22\columnwidth}
\begin{center}
\begin{tabular}{|cc|}
\hline \multicolumn{2}{|c|}{$nrngnb$}\\ \hline\multicolumn{2}{|c|}{} \\
\multicolumn{2}{|c|}{ $nr = \frac{R}{R+G+B}$ }\\
 \multicolumn{2}{|c|}{ $ng =
\frac{G}{R+G+B}$ } \\ \multicolumn{2}{|c|}{ $nb = \frac{B}{R+G+B}$
}\\\multicolumn{2}{|c|}{} \\\hline
\end{tabular}
\end{center}
\end{minipage}
\hspace{.5cm}
\begin{minipage}[b]{0.6\columnwidth}
\begin{center}
\begin{tabular}{|c|}
\hline \multicolumn{1}{|c|}{Opponent Color Space}\\ \hline \multicolumn{1}{|c|}{}\\
\multicolumn{1}{|c|}{ \hspace{-0.15cm}$\hspace{-0.05cm}\left(
\begin{array}{c} O_1 \\ O_2 \\ O_3 \end{array} \right)\hspace{-0.15cm} = \hspace{-0.15cm}\left(
\begin{array}{ccc}
\frac{1}{\sqrt{2}} & \frac{-1}{\sqrt{2}} & 0 \\
\frac{1}{\sqrt{6}} & \frac{1}{\sqrt{6}} & \frac{-2}{\sqrt{6}} \\
\frac{1}{\sqrt{3}} & \frac{1}{\sqrt{3}} &
\frac{1}{\sqrt{3}}\end{array} \right)\hspace{-0.15cm}\left  ( \begin{array}{c} R \\
G \\ B \end{array} \right)\hspace{-0.15cm}$ }\\ \multicolumn{1}{|c|}{}\\ \hline
\end{tabular}
\end{center}
\end{minipage}
\\
\begin{minipage}[b]{\columnwidth}
\begin{center}
\begin{tabular}{|cc|}
\hline \multicolumn{2}{|c|}{$HSV$}\\ \hline \multicolumn{2}{|c|}{
$\left(
\begin{array}{c} V \\ V_1 \\ V_2 \end{array} \right) = \left(
\begin{array}{ccc}
\frac{1}{3} & \frac{1}{3} & \frac{1}{3} \\
\frac{-1}{\sqrt{6}} & \frac{-1}{\sqrt{6}} & \frac{2}{\sqrt{6}} \\
\frac{1}{\sqrt{6}} & \frac{-2}{\sqrt{6}} &
\frac{1}{\sqrt{6}}\end{array} \right)\left( \begin{array}{c} R \\
G \\ B \end{array} \right)$ }\\ \multicolumn{2}{|c|}{}\\ $H
=\arctan{\frac{V_2}{V_1}}$ & $S = \sqrt{V_1^2 + V_2^2}$\\ \hline
\end{tabular}
\end{center}
\end{minipage}
\\
\begin{minipage}[b]{\columnwidth}
\begin{center}
\begin{tabular}{|cc|}
\hline \multicolumn{2}{|c|}{CIE $Lab$}\\ \hline
\multicolumn{2}{|c|}{ $\left(
\begin{array}{c} X \\ Y \\ Z \end{array} \right) = \left(
\begin{array}{ccc}
$0.490$ & $0.310$ & $0.200$ \\
$0.177$ & $0.812$ & $0.011$ \\
$0.000$ & $0.010$ & $0.990$ \end{array} \right)\left( \begin{array}{c} R \\
G \\ B \end{array} \right)$ }\\
\multicolumn{2}{|c|}{}\\\multicolumn{2}{|c|}{$L =116(\frac{Y}{Y_0})^{\frac{1}{3}}-16$} \\

$a = 500\left[
(\frac{X}{X_0})^{\frac{1}{3}}-(\frac{Y}{Y_0})^{\frac{1}{3}}\right]$
& $b = 200\left[
(\frac{Y}{Y_0})^{\frac{1}{3}}-(\frac{Z}{Z_0})^{\frac{1}{3}}\right]$\\\multicolumn{2}{|c|}{\scriptsize{$X_0$,
$Y_0$ and $Z_0$ are the coordinates of a reference white
point.}}\\
 \hline
\end{tabular}
\end{center}
\end{minipage}
}
\end{table}
\subsection{Pixel Classification}
\label{subsect:classification} The second stage of the algorithmic
pipeline takes converted pixel values as input and outputs a
pixel-level road likelihood based on one-class classification.
One-class classification corresponds to the problem of
distinguishing the target class from all other possible classes
which are considered as non-targets or outliers. We assume that
only examples of the target class are available for training. This
is because it is assumed that non-target samples are not present
or not properly sampled. In fact, binary classifiers relying on
training samples from both classes are not considered as they can
not create a boundary between the two classes during the training
process. We consider one class classifiers as they characterize
the target class and then, given a test sample, decide whether it
belongs or not to that class. As a consequence, one class
classifiers assume that a well-sampled training set of the target
objects is available. Ideally, the model description of the target
class should be large enough to accept most of the new target
samples and yet selective to reject outliers. However, in online
road detection, collecting road samples is an ill-posed problem
since the knowledge of the road class is deduced from a finite
(and small) set of training samples (in our case, collected in a
unsupervised manner from the bottom part of the image). Hence, the
additional problem arises of having a poorly sampled target class
(we do not have a sufficient number of samples of the target
class) leading to ill-posed representations and distributions.

\begin{table}[t!]
\caption{Properties of different color representations from a theoretical
point of view. Robustness to each property is indicated by '+' and weakness by '-'.} \label{tab:taxonomy}
\begin{center}
\begin{tabular}{|p{1.2cm}|p{0.15cm}|p{0.15cm}|p{0.15cm}|p{0.15cm}|p{0.15cm}|p{0.15cm}|p{0.15cm}|p{0.15cm}|p{0.15cm}|p{0.15cm}|p{0.15cm}|p{0.15cm}|}\cline{2-12}
\multicolumn{1}{p{1.2cm}|}{\centering \scriptsize{}} &  \multicolumn{1}{p{0.15cm}|}{\begin{sideways}\scriptsize{$R$}\end{sideways} } & \multicolumn{1}{p{0.15cm}|}{\begin{sideways}\scriptsize{$G$}\end{sideways} } &\multicolumn{1}{p{0.15cm}|}{\begin{sideways}\scriptsize{$B$}\end{sideways} } &\multicolumn{1}{p{0.15cm}|}{\begin{sideways}\scriptsize{$nr$}\end{sideways} } & \multicolumn{1}{p{0.15cm}|}{\begin{sideways}\scriptsize{$ng$}\end{sideways}} & \multicolumn{1}{p{0.15cm}|}{\begin{sideways}\scriptsize{$O_1$}\end{sideways}} & \multicolumn{1}{p{0.15cm}|}{\begin{sideways}\scriptsize{$O_2$}\end{sideways}} &  \multicolumn{1}{p{0.15cm}|}{\begin{sideways}\scriptsize{$L$, $V$}\end{sideways}}  &  \multicolumn{1}{p{0.15cm}|}{\begin{sideways}\scriptsize{$ab$}\end{sideways}} &  \multicolumn{1}{p{0.15cm}|}{\begin{sideways}\scriptsize{$H$}\end{sideways}} &  \multicolumn{1}{p{0.15cm}|}{\begin{sideways}\scriptsize{$S$}\end{sideways}} \\
 \hline
 \multicolumn{1}{|p{1.2cm}|}{\centering \scriptsize{Global Illumination changes}} &
 - & - & - & - & - & + & + & - & + & + &  + \\\hline
 \multicolumn{1}{|p{1.2cm}|}{\centering \scriptsize{Discriminative power}} &
 + & + & - & - & + & - & - & - & + & - &  - \\\hline
 \multicolumn{1}{|p{1.2cm}|}{\centering \scriptsize{Distance to the camera}} &
 + & + & + & + & + & + & + & + & + & + &  + \\ \hline
 \multicolumn{1}{|p{1.2cm}|}{\centering \scriptsize{Strong shadows}} &
 - & + & - & - & - & + & + & - & + & + &  + \\ \hline
 \multicolumn{1}{|p{1.2cm}|}{\centering \scriptsize{Highlights}} &
 - & - & - & - & - & + & + & - & + & + & + \\ \hline
 \multicolumn{1}{|p{1.2cm}|}{\centering \scriptsize{Road type and shape}} &
 + & + & + & + & + & + &  + & + & + & + & + \\ \hline
 \end{tabular}
\end{center}
\end{table}

\begin{figure*}[t!]
\begin{center}
\begin{tabular}{ccc}
\includegraphics[width=0.22\textwidth]{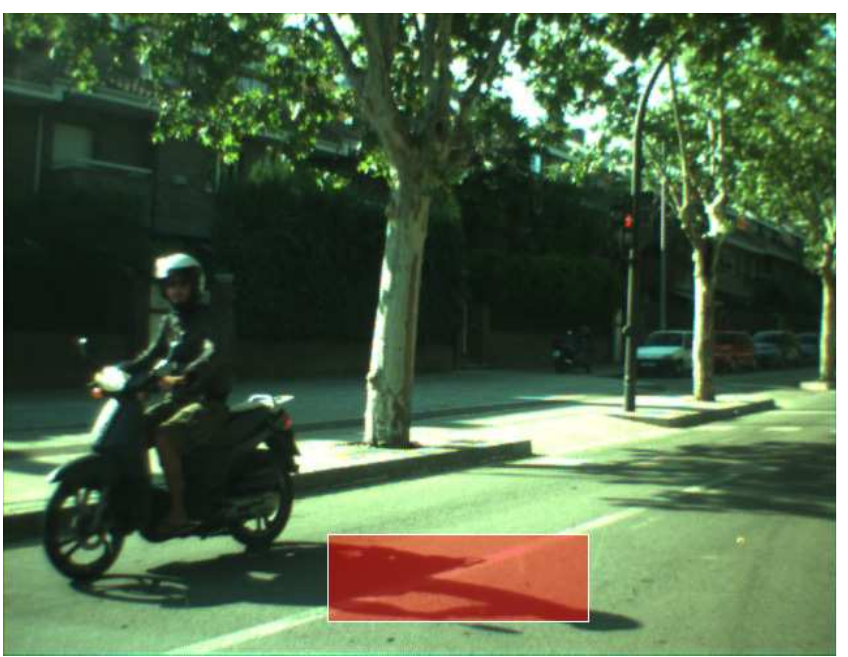}&
\includegraphics[width=0.22\textwidth]{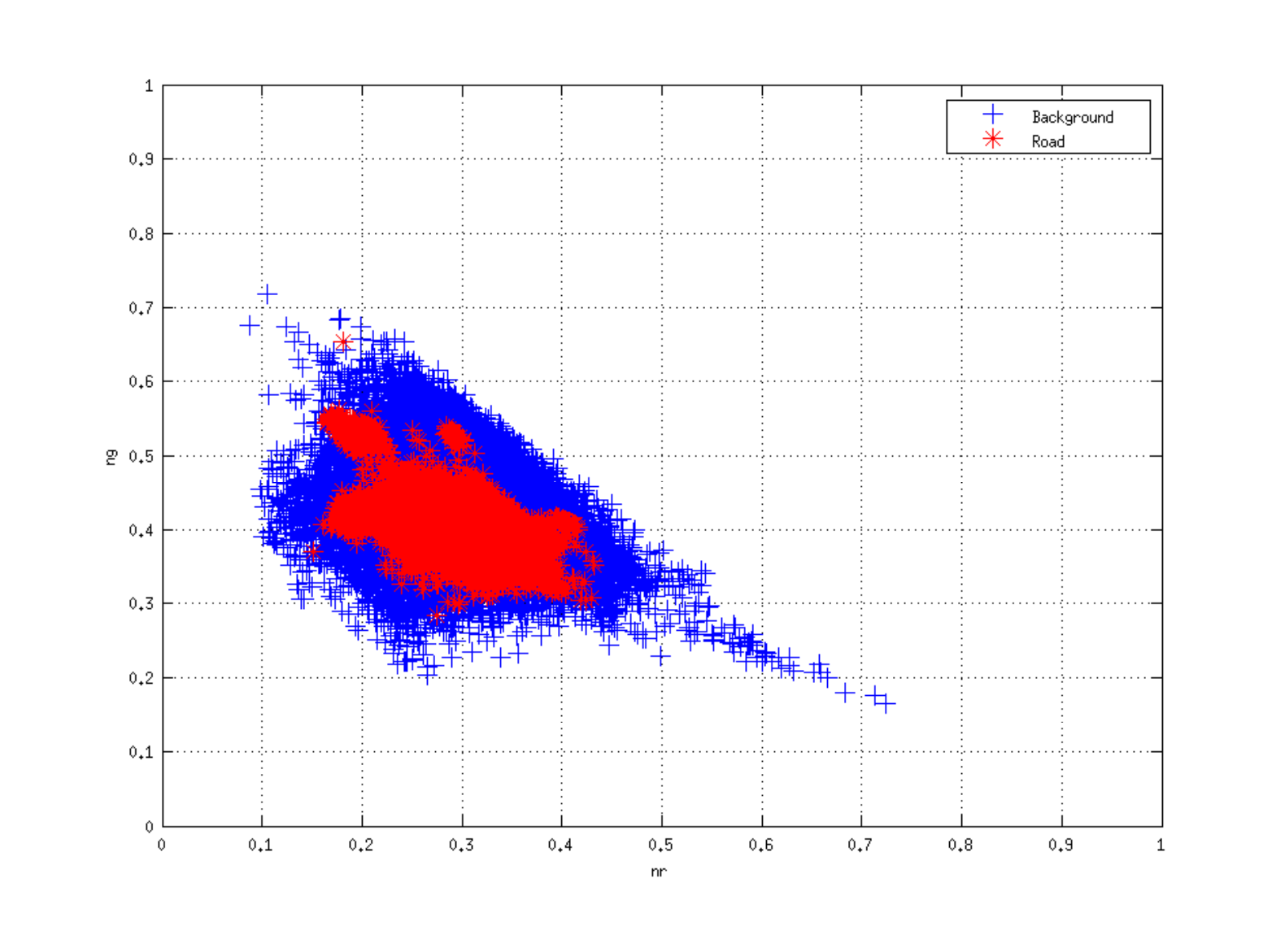}&
\includegraphics[width=0.22\textwidth]{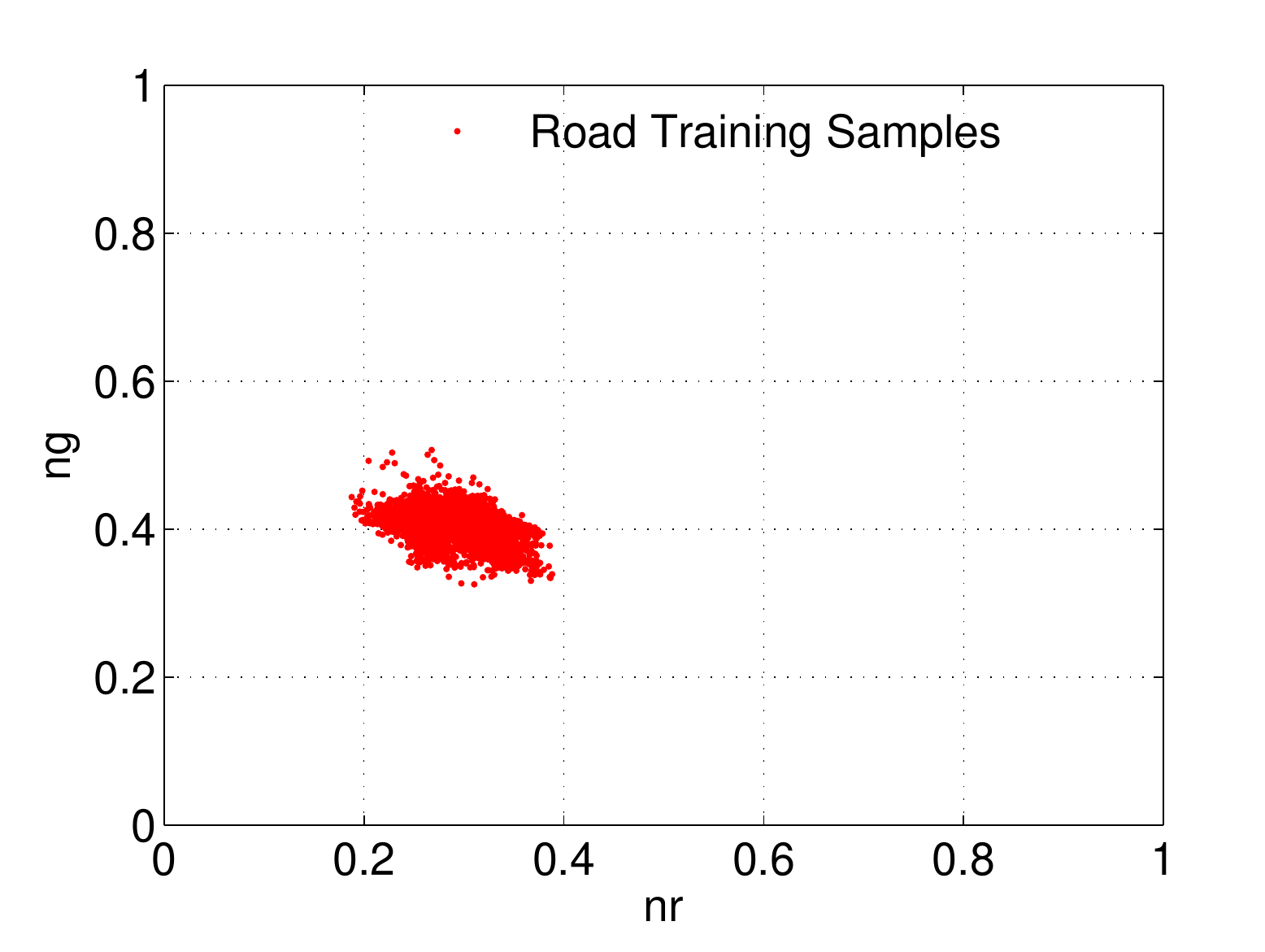}\\
(a)&(b)&(c)\\
\includegraphics[width=0.22\textwidth]{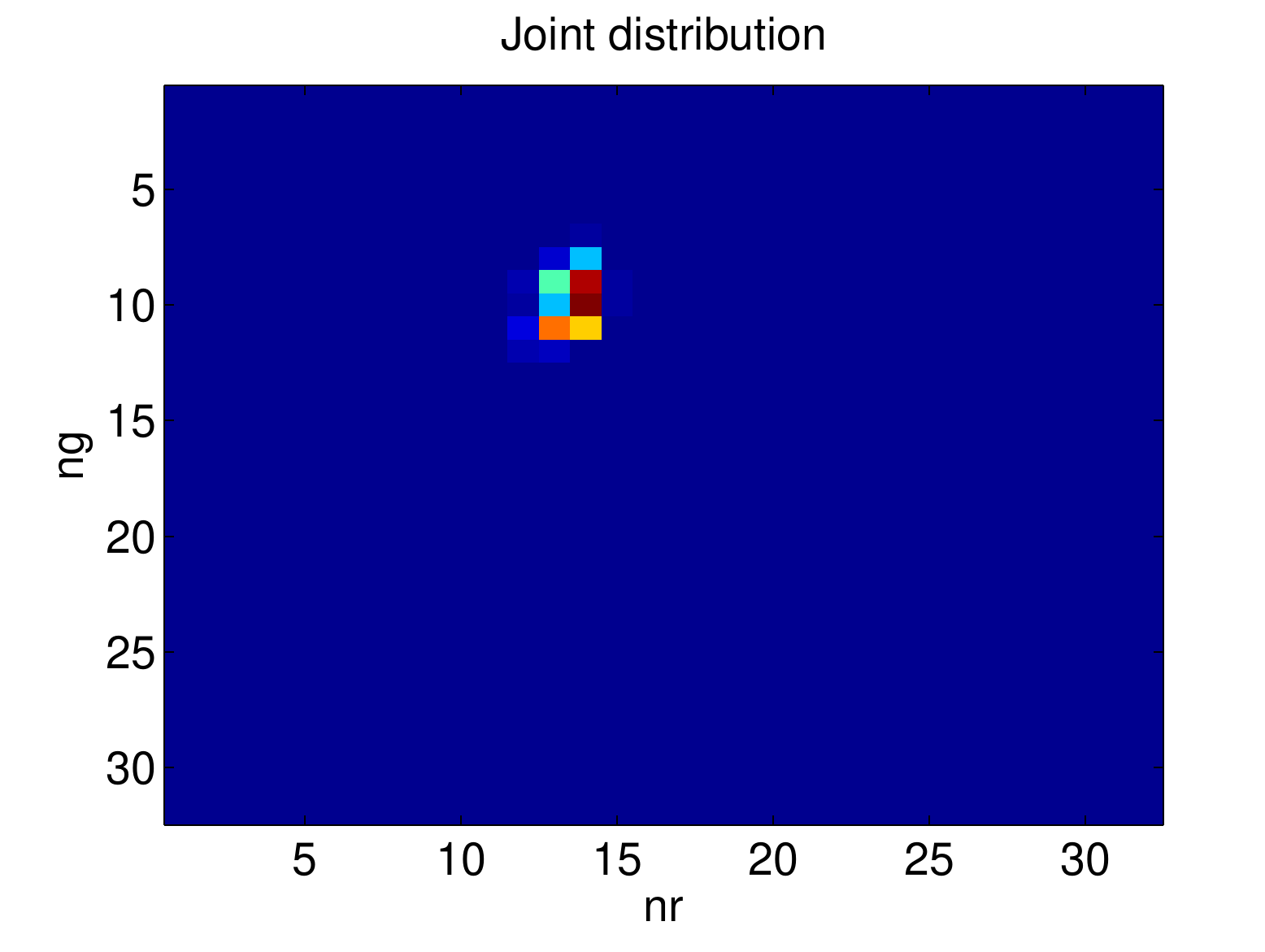}&
\includegraphics[width=0.22\textwidth]{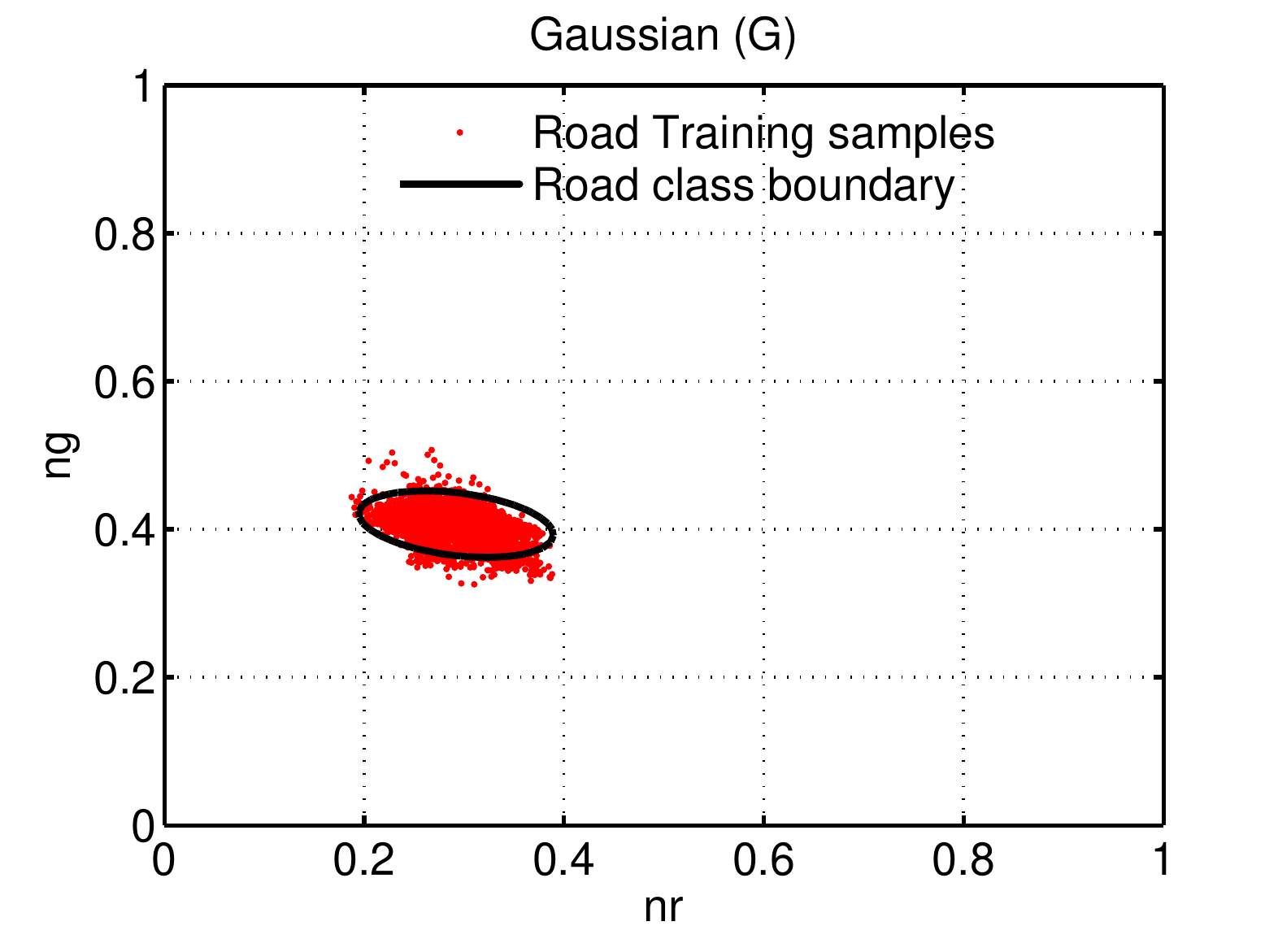}&
\includegraphics[width=0.22\textwidth]{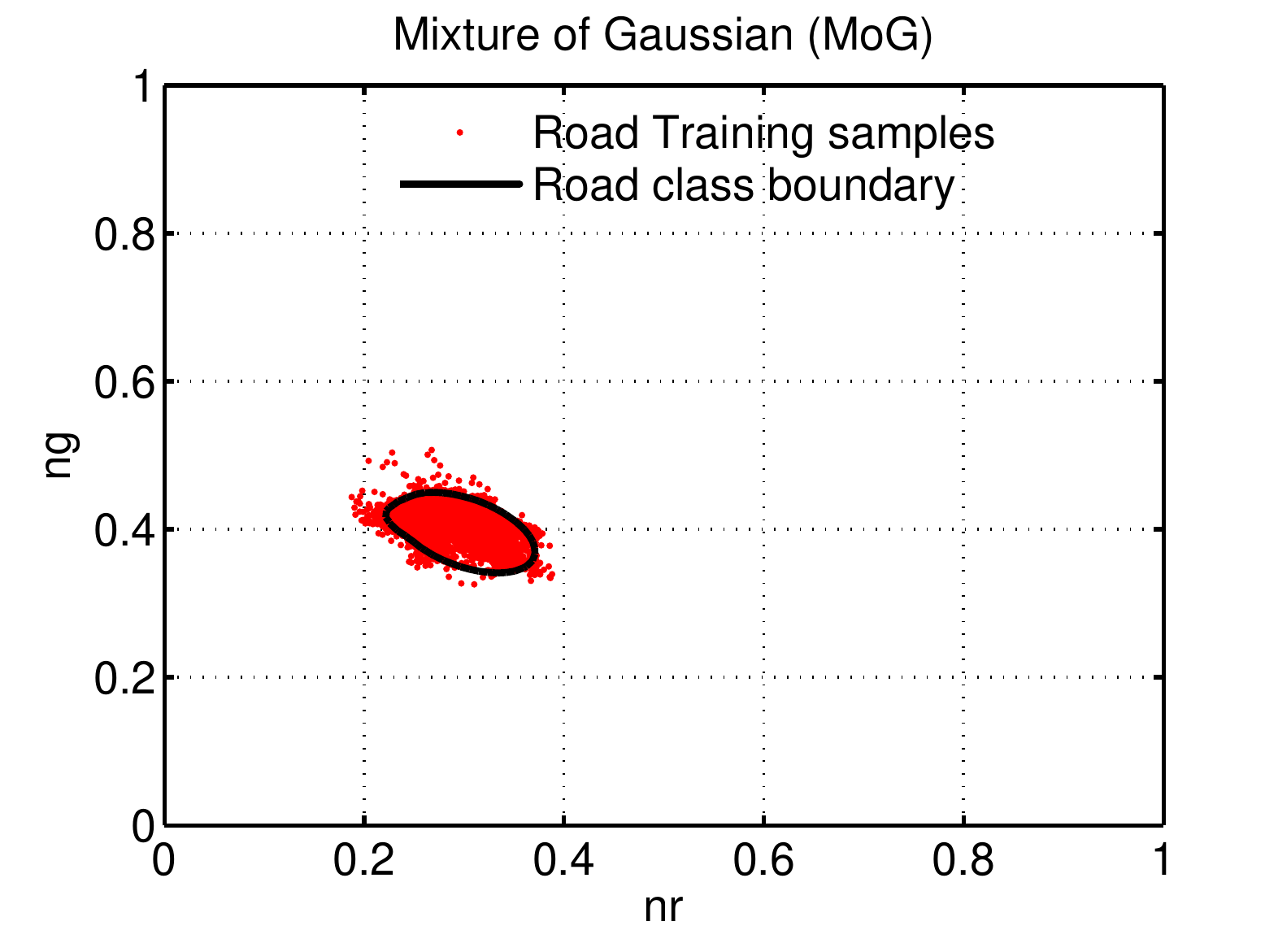}\\
(d)&(e)&(f)\\
\end{tabular}
\begin{tabular}{cccc}
\includegraphics[width=0.22\textwidth]{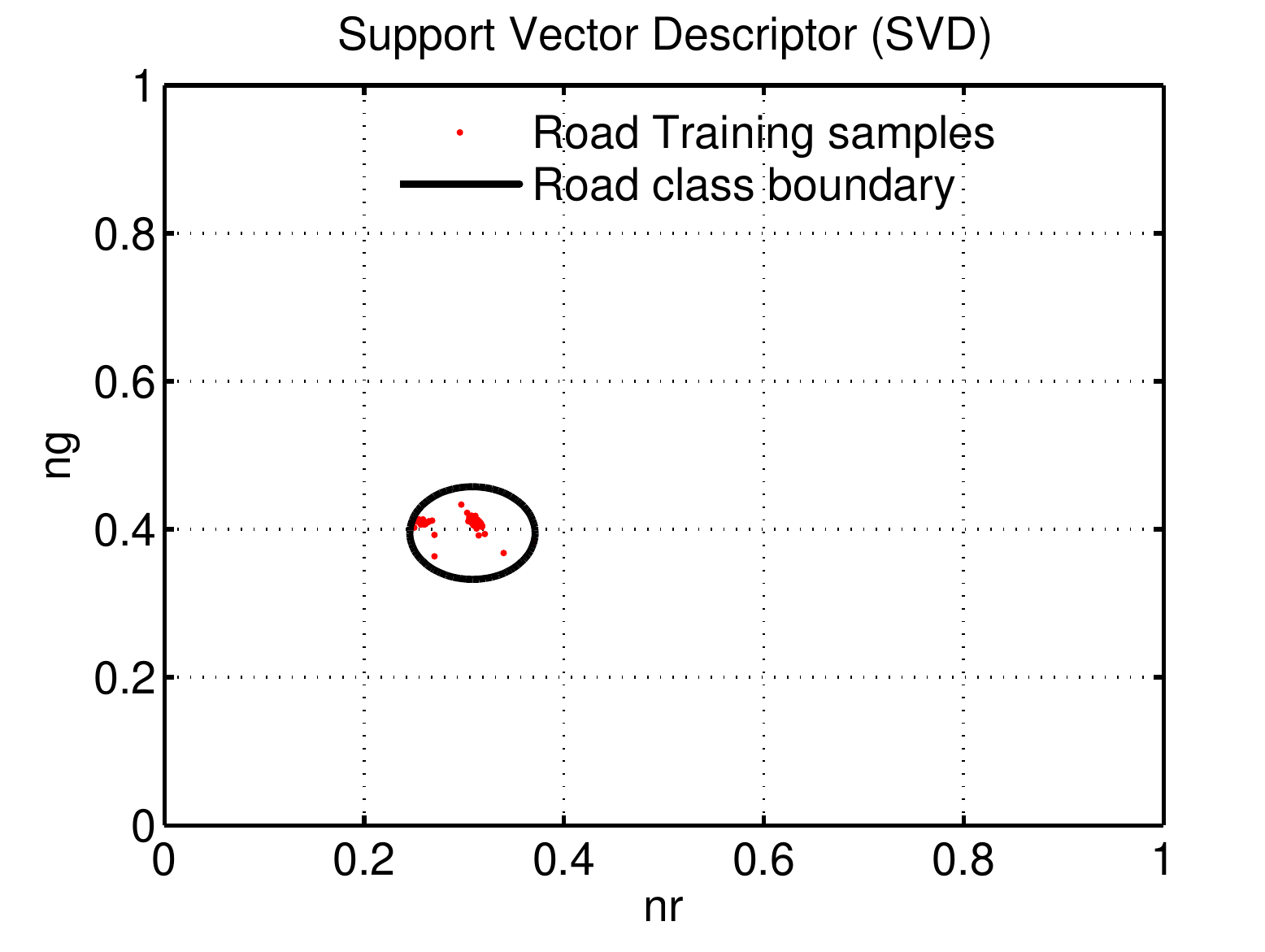}&
\includegraphics[width=0.22\textwidth]{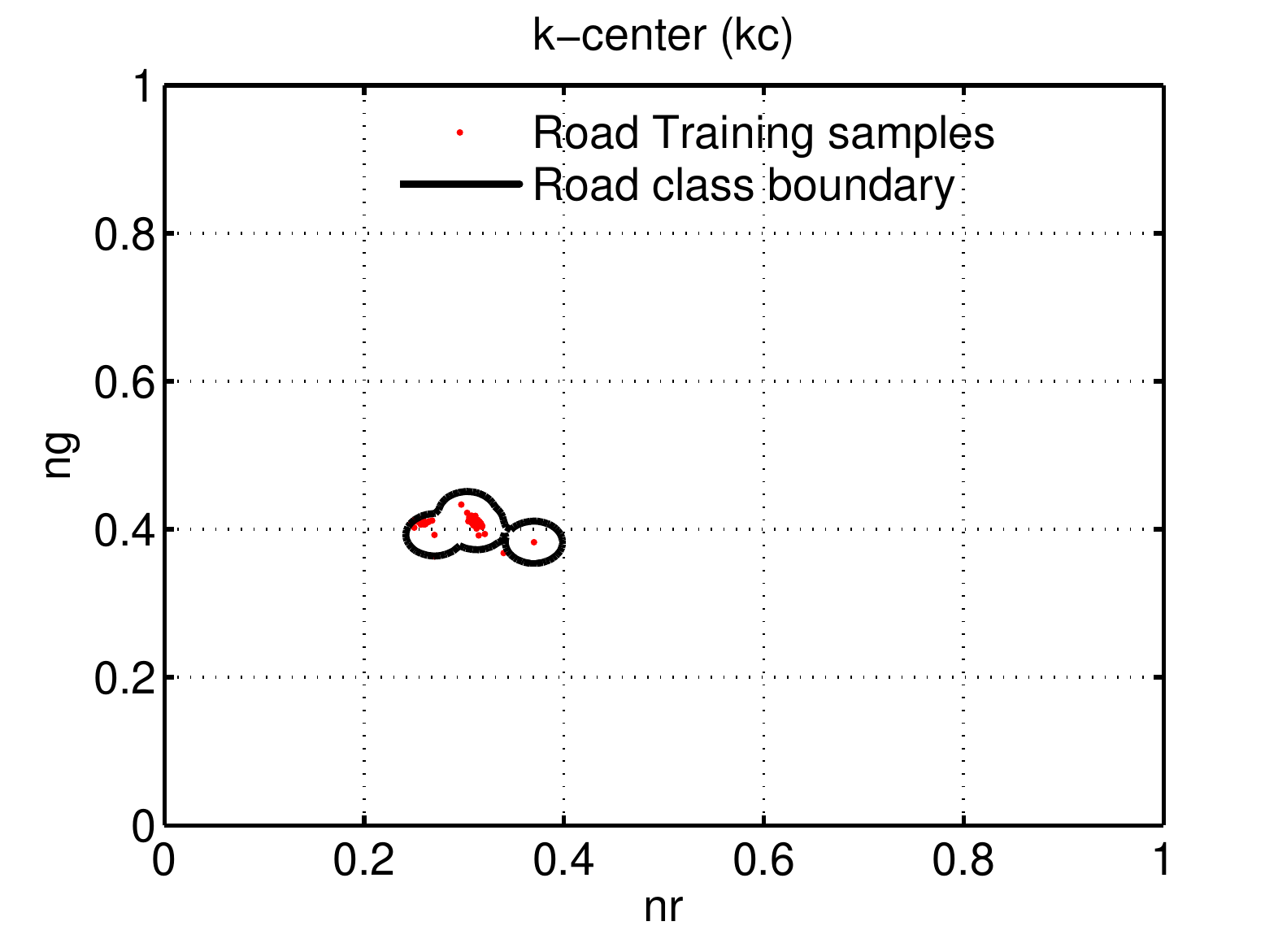}&
\includegraphics[width=0.22\textwidth]{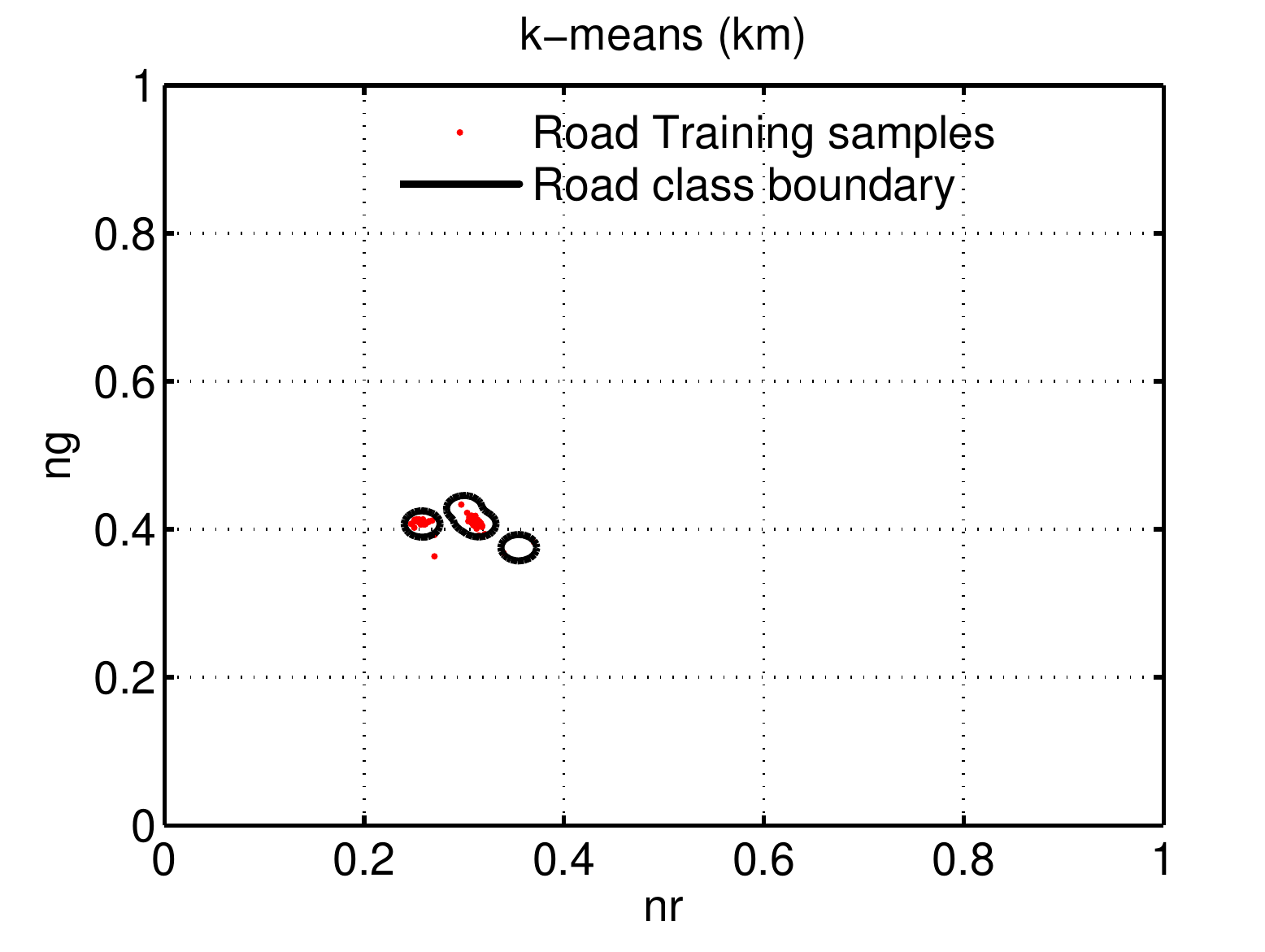}&
\includegraphics[width=0.22\textwidth]{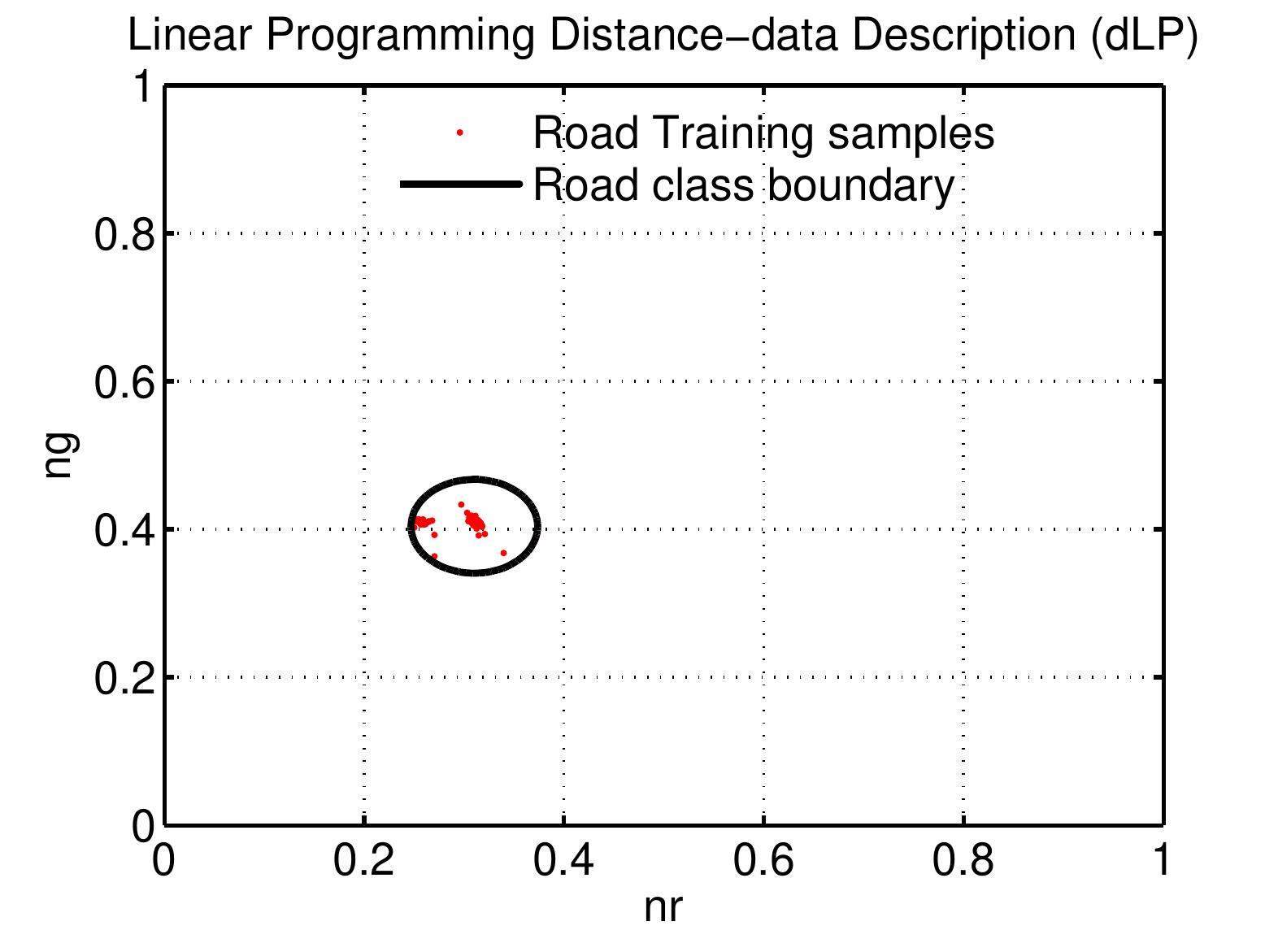}\\
(g)&(h)&(i)&(j)\\
\end{tabular}
\end{center}
\caption{a) Input image showing the ROI area used for training. b)
Distribution of road and background pixels in normalized $RG$
color space ($nr$,$ng$). c) Distribution of road pixels within the
training area in the same color space. d) Joint distribution of
the training data. e) and f) Gaussian and Mixture of Gaussian
representations of training data. Bottom row shows classifiers
based on superpixels for training: g) Support vector descriptor;
h) k-centers; i) k-means and j) Linear programming data
descriptor. As shown, using the centroid of superpixels reduces
the variance in the input data.  Detailed description of these
classifiers can be found in~\sect{subsect:classification}.
}\label{fig:oneclassExample}
\end{figure*}

One-class classifiers can be divided in three
groups~\cite{TaxPhD:2001}: density based, reconstruction based and
boundary methods. Density based methods aim at modeling the
probability density function of the target class using training
data. Reconstruction and boundary based methods avoid the explicit
estimation of the probability density function. The former is
based on assumptions of the underlying structure of the data. The
latter aims at defining the boundaries that encloses all the
elements from the target class (in the training set).

In the rest of this section, we briefly review most of the
promising one-class classification algorithms. First, we focus on
five different density methods: model-based (histograms),
nearest-neighbors, single Gaussian, robustified Gaussian and
mixture of Gaussians. Then, two reconstruction methods are
discussed such as the $k$-means and Principal Component Analysis
algorithms. Finally, seven boundary methods are outlined:
nearest-neighbor, k-centers, linear data description, support
vector description, min-max probability and minimum spanning tree
method. The evolution of these methods for a given road image is
shown in~\fig{fig:oneclassExample}

\textbf{Model-based (Mb)}. This is a non-parametric classifier
that uses a likelihood measure to approximate the conditional
probability of having a road pixel given a pixel value. This
probability distribution is estimated for each image using the
training samples. In particular, we use the normalized histogram
of training samples. Therefore, the road likelihood is given by
$\mathcal{L}_{i} = p(x_i)$, where $p(\cdot)$ is the normalized
histogram. The higher the likelihood value, the higher the
potential of being a road pixel.

\textbf{Single Gaussian (G)}. This classifier models road training
samples using a unique Gaussian distribution. The road likelihood
for the i-th pixel is obtained by
$\mathcal{L}_i=G(x,\mu_r,\sigma_r)$, where $x_i$ is the pixel
value and $\mu_r$, $\sigma_r$ are the parameters of the Gaussian
distribution learned using the training samples. In practice, to
avoid numerical instabilities, we do not estimate the density.
Instead, we use the Mahalanobis distance as follows:
$\mathcal{L}_i=(x_i-\mu_r)^T\Sigma_r^{-1}(x_i-\mu_r)$, where
$\Sigma_r$ is the covariance matrix estimated using the training
set.

\begin{figure*}[t!]
\begin{center}
\includegraphics[width=\textwidth]{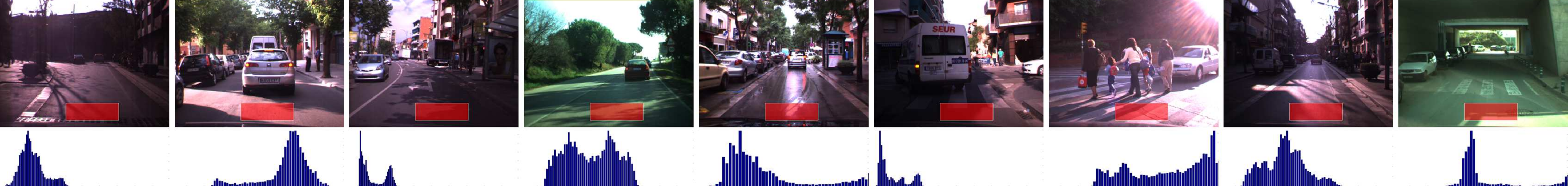}
\end{center}
\caption{Top row shows examples of test images with the training area overlaid in red. The bottom row shows the gray-level value distribution for each training area. As shown, these distributions vary considerably depending on the acquisition conditions.}\label{fig:distributionexamples}
\end{figure*}

\textbf{Robustified Gaussian (RG)}. The single Gaussian classifier
is sensitive to outliers and noise in the training samples. In our
case, these outliers are long tails in the distribution mainly due
to lighting conditions or different road appearances as shown
in~\fig{fig:distributionexamples}. Therefore, the robustified
Gaussian classifier is based on a single Gaussian where the
parameters are learned using robust statistics. To achieve this,
training samples are weighted according to their proximity to the
mean value. Distant samples are down weighted to obtain a more
robust estimate. Finally, the road likelihood is obtained as in
Gaussian classifier.

\textbf{Mixture of Gaussians (MoG)}. Single Gaussian classifiers
have the drawback of modeling a single distribution. This may
negatively influence their performance in the presence of shadows
and lighting variations. Mixture of Gaussians classifier models
the set of training samples using a combination of $N$ Gaussians
and thus, creates a more flexible description of the road class.
The road likelihood is given by $\mathcal{L}_i = \sum_{n=1}^N P_n
exp^{-(x-\mu_n)^T\Sigma_n^{-1}(x_i-\mu_n)}$, where $\mu_n$ and
$\Sigma_n$ are the parameters of the different Gaussians involved
and $P_n$ is the weight assigned to the n-th Gaussian. In this
paper, we optimize these parameters using the EM algorithm and we
will also evaluate different values of $N$.

\textbf{k-means (km)}. This classifier does not rely on estimating
the density probability function. Instead, the classier describes
the training data using $k$ different clusters. These clusters are
defined by minimizing the average distance to a cluster center.
Then, the road likelihood is obtained by $\mathcal{L}_i= min_j
(||x_i-c_j||^2)$ where $\mathbf{c}=[c_1,\dots,c_k]$ is the set of
cluster centers.

\textbf{k-center (kc)}. This method aims at covering the training
set with $k$ small balls with equal radious. The centers of these
balls $c_1,\dots,c_k$ are placed on training samples by minimizing
the maximum distance of all minimum distance between training
pixels and the centers of the balls (minimize
$max_j(min_k||y_i-c_k||^2)$). Once the centers are defined, the
road likelihood is obtained as in the $k$-means method:
$\mathcal{L}_i= min_j (||x_i-c_j||^2)$.
%
%

\textbf{Principal Component Analysis (PCA)}. This classifier
describes road data using a linear subspace defined by the
eigenvectors of the data covariance matrix. To verify if a new
data instance belongs to the road class, the algorithm analyzes
the reconstruction error defined as the difference between the
incoming instance and the projection of that instance in the road
subspace. Therefore, the road likelihood is defined by
$\mathcal{L}_i=||x_i-x_i^{proj}||^2$, where $x_i^{proj}$ is the
projection of $x_i$ into the subspace. In this paper, we assume
the subspace is built using the eigenvectors with the largest
eigenvalues representing $95\%$ of the energy in the original
data.

\textbf{Nearest neighbor (NN)}. This method avoids the explicit
density estimation and estimates the road likelihood using the
distances between test pixels and the training data. That is,
$\mathcal{L}_i= min_j (x_i-y_j)^2$ where
$\mathbf{y}=[y_1,\dots,y_N]$ is the set of $N$ training pixels. In
this paper, we consider the minimum squared Euclidean distance
over the training set. However, this method is suitable to use any
other metric such as circular distances over specific color
planes.

\textbf{Linear programming distance-data description (dLP)}. This
method aims at describing the road data in terms of distances to
other objects~\cite{PekalskaTD02}. Then, the road likelihood is
estimated based on the dissimilarity between the test pixels and
road training samples. This is formulated using a linear proximity
function as follows: $\mathcal{L}_i= \sum_i{w_id(x,y_i)}$, where
the weights $w$ are optimized to minimize the max-norm distance
from the bounding hyperplane to the origin. Furthermore, only a
few of these weights are non-zero as a consequence of the linear
programming formulation.

\textbf{Support Vector Descriptor (SVD)}. This method aims to
define the hypersphere with a minimum volume covering the entire
training set~\cite{Tax1999c}. This is a specific instance of the
SVM classifier where only positive examples are used. In our case,
we consider a general kernel to fit a hypersphere around the road
samples in the training set. Then, the road likelihood is computed
as the distance of the test sample to the center of the sphere.

\textbf{Minimax Probability (MPM)} This method aims at computing
the linear classifier that separates the data from the origin
rejecting maximally a specific fraction of the training data
represented as a random variable
$y\sim(\bar{y},\Sigma_y)$~\cite{Lanckriet:2003}.

\textbf{Minimum Spanning Tree (MST)}. This is a non-parametric
classifier aiming at capturing the underlying structure of the
data based on fitting a minimum spanning tree to the training
data~\cite{Juszczak:2009}. In the ideal case, a test instance
belongs to the target class if it is in one of the edges of the
spanning tree. However, since the training set is finite and may
not represent all possible instances of the target class, a test
instance is considered as a target if it lies in the neighborhood
of any of the edges. Therefore, the road likelihood is estimated
as the minimum distance to the one of the edges of the tree given
by $\mathcal{L}_i= ||x_i-p_{e_{lj}}(x_i)||$, where $p_{e_{lj}}(x)$
is the projection of the test pixel $x_i$ onto the line defined by
two training samples $\{y_l,y_j\}$ (\ie, the vertices of the
tree). In those cases where the projection does not lie between
$y_l$ and $y_j$ then, the distance is computed as a nearest
neighbour distance between $x_p$ and $y_l$ or $y_j$.
 \begin{figure}[t!]
 \begin{center}
 \includegraphics[width=\columnwidth]{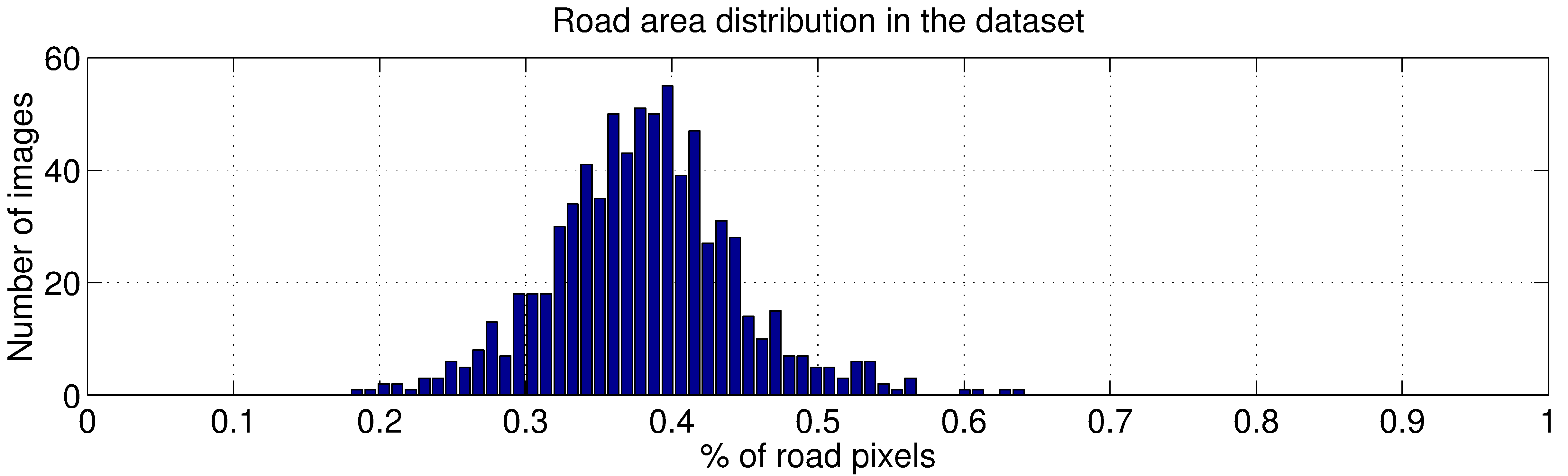}
 \end{center}
 \caption{Distribution of images according to the percentatge of road occupancy. As shown, the dataset consists of images where at most $35-40\%$ of the image belongs to the road surface.}
 \label{fig:BBDDdistribution}
 \end{figure}
 
\begin{figure*}[t!]
\begin{center}
\begin{tabular}{cccc}
\hspace{-0.3cm}\includegraphics[width=0.33\textwidth,height = 14cm]{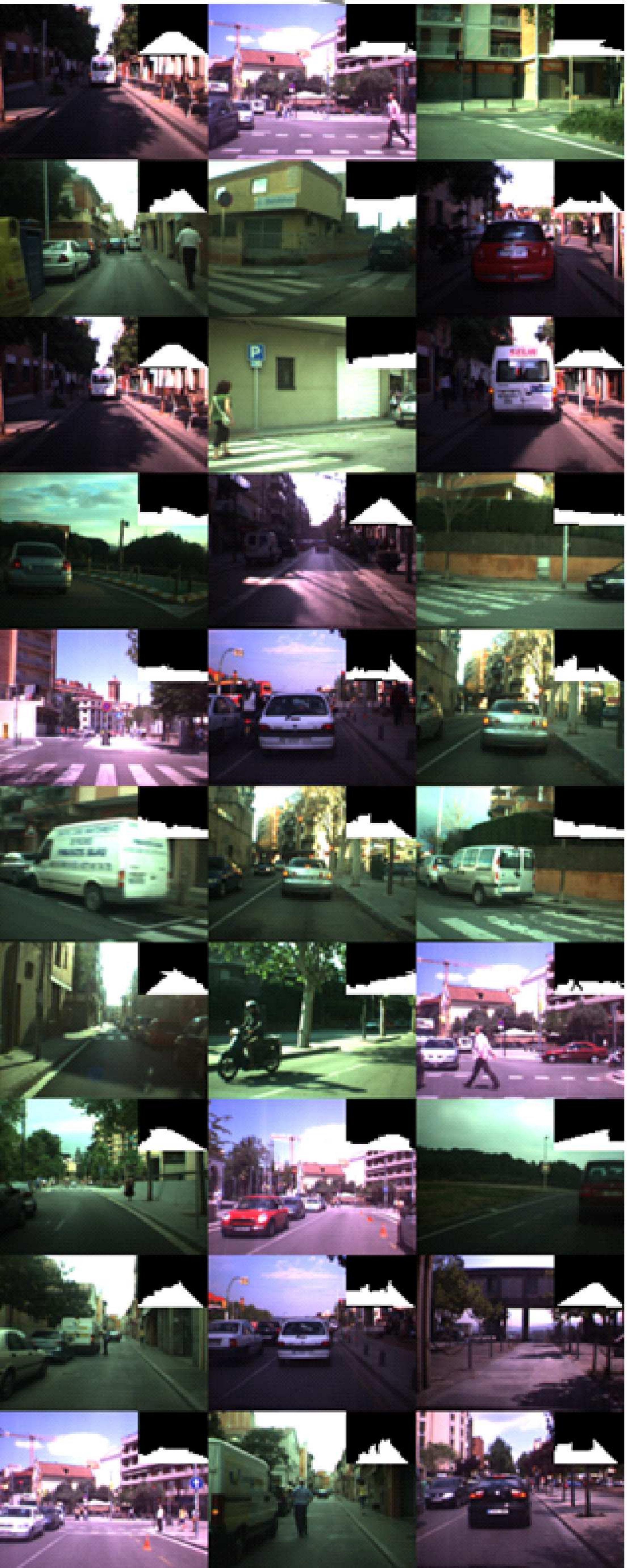}&
\hspace{-0.3cm}\includegraphics[width=0.33\textwidth,height = 14cm]{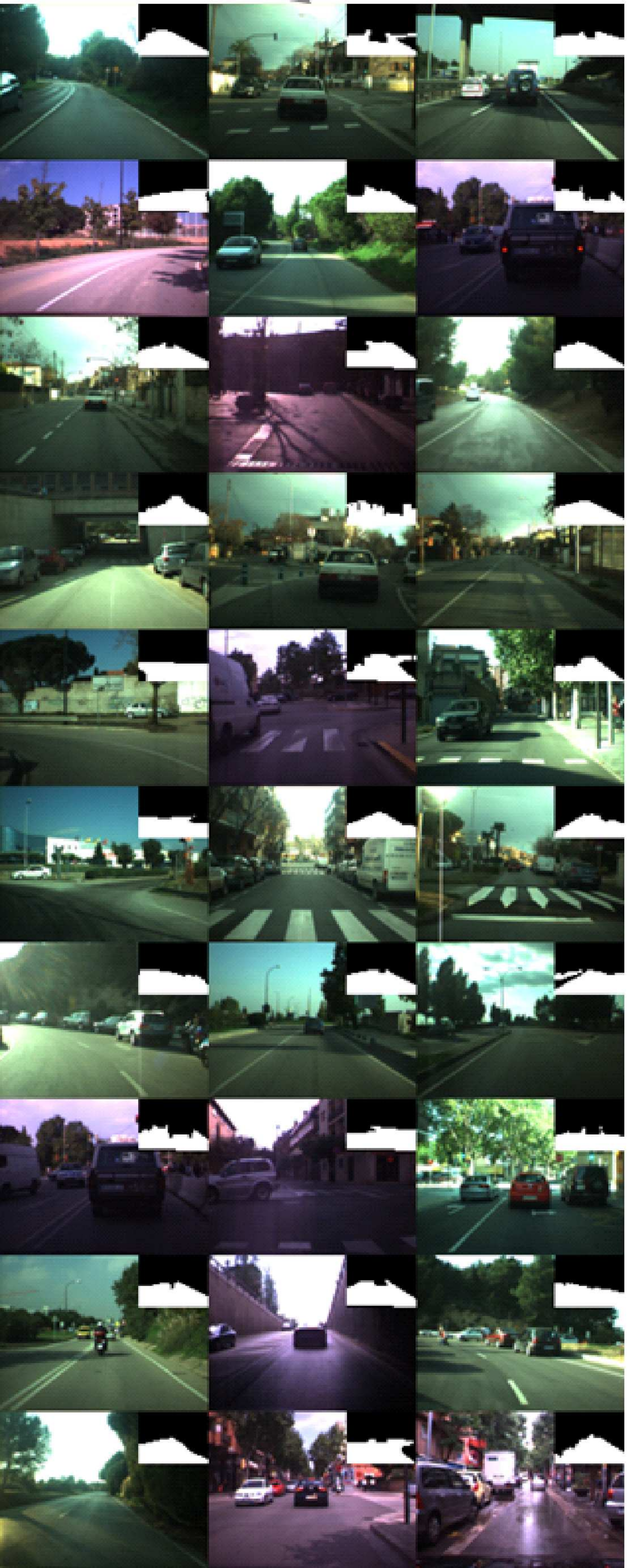}&
\hspace{-0.3cm}\includegraphics[width=0.22\textwidth,height = 14cm]{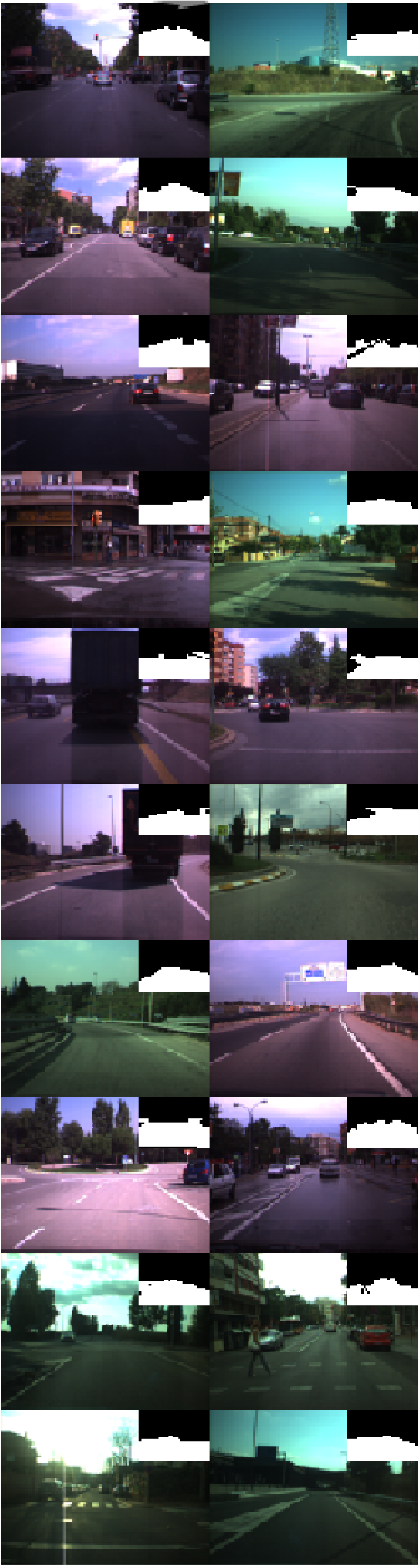}&
\hspace{-0.3cm}\includegraphics[width=0.109\textwidth,height = 14cm]{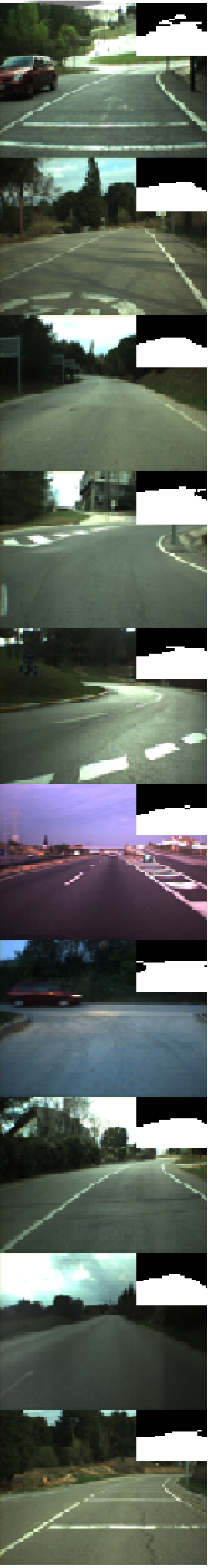}\\
(a)&(b)&(c)&(d)\\
\end{tabular}
\end{center}
\caption{Example of annotated images in the dataset. (a) Images showing at most $30\%$ of road areas; (b) Images showing between $30\%$ and $45\%$ of road areas; (c) Images showing between $45\%$ and $55\%$ and (d) images showing more than $55\%$ of road areas.}
\label{fig:BBDD} 
\end{figure*}

\section{The Road Dataset}
\label{sect:BBDD}
In this section, we introduce and provide a novel dataset for road
detection. The dataset consists of $755$ still images extracted
from different road sequences comprising thousands of images
acquired at different days, different daytime (daybreak, morning,
noon and afternoon), different weather conditions (sunny, cloudy,
rainy) and mainly from urban-like scenarios. The set of images has
been carefully selected to include the major challenges in real
world driving situations by discarding those images where the road
is uniformly illuminated. We also discard those images where the percentatge of the image covered by the road surface is too large leading to the distribution of images shown in~\fig{fig:BBDDdistribution}. As shown, the dataset consists of images where the road represent approximately the $45\%$ of the image. Images in the dataset contain strong shadows, wet
surfaces, sidewalks similar to the road, direct reflections,
crowded scenes and lack of lane markings as shown~\fig{fig:BBDD}.

\begin{figure}[t!]
\begin{center}
\includegraphics[width=\columnwidth]{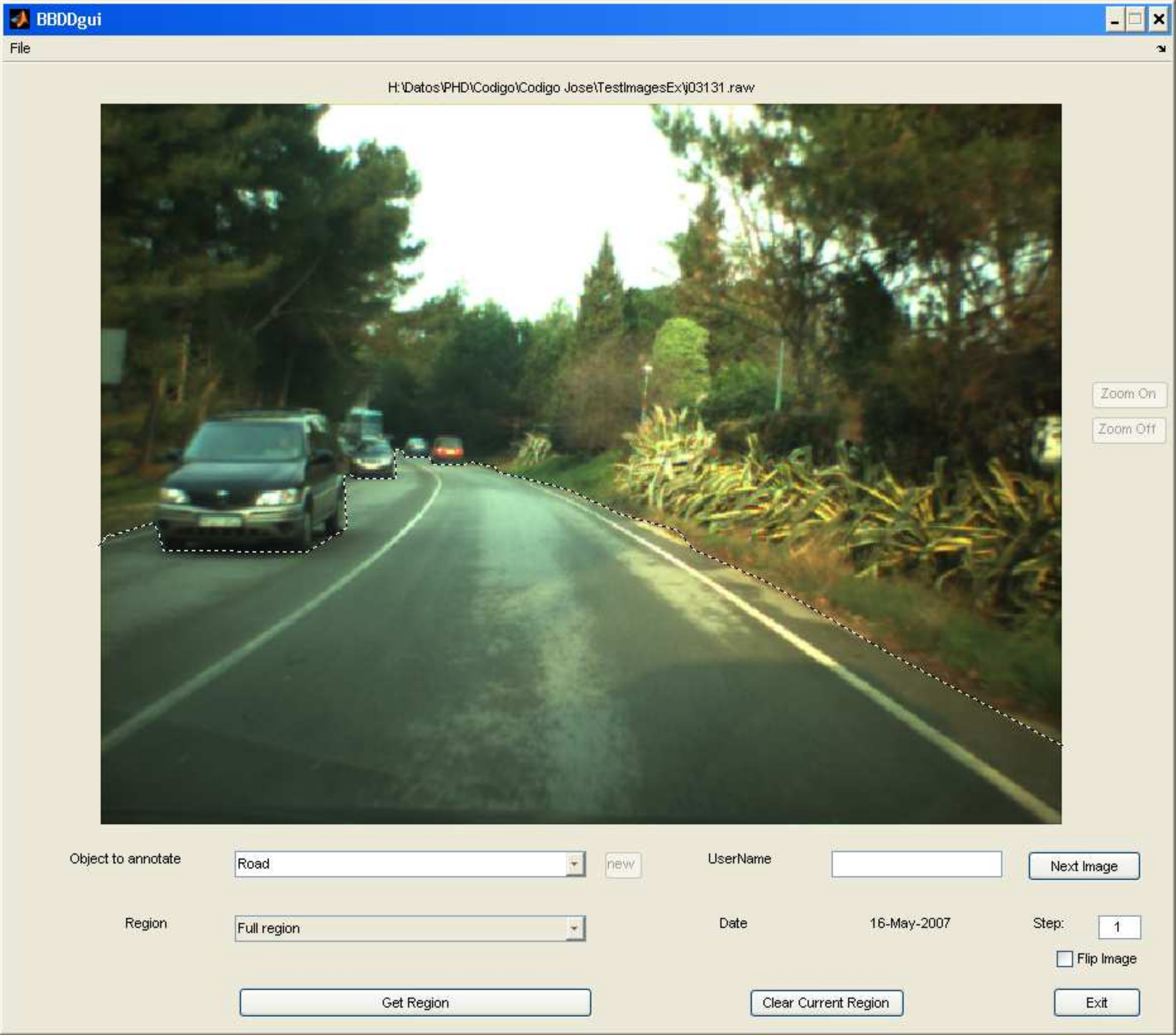}
\end{center}
\caption{User interface used to visualize and annotate road based
images. The application enables the user to label
different types of objects.
Moreover, each object can be labeled by different users and
composed by different blobs. The final file is created using XML
notation.} \label{figure:BBDDgui}
\end{figure}

Ground-truth is provided by a single experienced user providing
manual segmentations~(\fig{fig:BBDD}). To facilitate the labeling
task, we used the annotation tool shown in~\fig{figure:BBDDgui}.
This tool allows multiple user annotations as well as defining
multiple objects such as cars, road, sky. Once the annotation is
completed, points defining the polygon around the object are
stored in a XML file associated to the user. Both the dataset and
annotation tool are made publicly available to the community at
http://scrd.josemalvarez.net.

\section{Experiments}
\label{sect:experiments}
In this section, we present experiments conducted to evaluate
different combinations of single class classifiers and color
representations for road detection. In particular, we evaluate
each color plane individually ($13$ color planes) and their most
common combinations such as $HSV$, $nrng$, $HS$, $O_1O_2$ and
$RGB$ in conjunction with a one-class classifier. The set up of
the classifiers is as follows. First, we consider four instances
of the model-based classifier. Two of these instances directly use
the training samples from the bottom part of the road to build the
normalized-histogram with $64$ and $100$ bins. The other two
instances extend the training set with noisy samples. Extending
the training set with synthetic samples is a common practice to
improve the robustness of the algorithms~\cite{AlvarezECCVW:2012}.
Hence, we duplicate the samples and adding zero mean and $30/256$
standard deviation noise to half of it (referred as $S\&N$). Then,
two different model-based configurations are considered: $100$ and
$64$ bins. Using different number of bins to build the histogram
enables the stability analysis of variations of this parameter.
The single and robustified Gaussian models are learned by
rejecting $2.5\%$ of the data. Furthermore, we consider three
instances of MoG classifier: $N=2$, $N=4$ and $N=opt$. This last
configuration optimizes $N$ based on the training set.

Road samples collected from a rectangular area ($201\times66$
pixels) at the bottom part of each image yields $13266$ training
pixels~(\fig{fig:distributionexamples}). Note that this area is
suited for right driving situations and it is not extremely large.
Furthermore, the area is fixed and independent of the image.
Therefore, as shown in \fig{fig:distributionexamples}, training
pixels may not represent all the road areas in the image for two
reasons: the variability within the training set is not
significant and the area does not belong to the road surface. To
reduce the computational cost required to train some methods, this
area is oversegmented using superpixels and only a single value
per superpixel is considered. In particular, we consider the
central value of the distribution within each super pixel to
reduce the effect of long tails due to noise in the imaging
process. This process reduces the training set to a compact area
of approximately $90$ samples per image as shown
in~\fig{fig:oneclassExample}.

\subsection{Evaluation Measures}

Quantitative evaluations are provided using average ROC
curves~\cite{Fawcett:2006} on the pixel-wise comparison between
ground-truth and results obtained by binarizing the road
likelihood $\mathcal{L}$ (\sect{sect:algorithm}) with different
threshold values. ROC curves represent the trade-off between true
positive rate $TPR$ and false positive rate
$FPR$~\cite{Alvarez:2008Ev}. These two measures provide different
insights into the performance of the algorithm. The true positive
rate ($TPR = \frac{TP}{TP+FN}$) or sensitivity refers to the
ability of the algorithm to detect road pixels. A low sensitivity
corresponds to under-segmented results.

False positive rate ($FPR = \frac{FP}{FP+TN}$) or fall-out refers
to the ability of the algorithm to detect background pixels.
Hence, a high fall-out corresponds to over-segmented results.
However, in road images, a low fall-out does not ensure a high
discriminative power since the number of false positives that can
appear within the road areas is negligible compared to the number
of background pixels. Hence, small fall-out variations may
correspond to significant variations in the final road detection
result. Finally, for performance comparison, we consider the area
under the curve (AUC $\in[0..1]$). The higher the AUC, the higher
the accuracy will be. The equal error rate (EER) is defined as the
intersection between the curve and the line where error rates are
equal \ie, $(1-TPR)=FPR$.
\begin{table*}[t!]
\caption{Quantitative results: a) Combinations of classifiers and single color planes. b) Combinations of classifiers and multiple color planes. The bold value indicates the best performance: a mixture of $2$ Gaussians in the $HSV$ color space.} \label{tab:evalNaming}
\begin{center}
{\small
\begin{tabular}{|p{0.1cm}|p{0.4cm}|p{0.3cm}|c|c|c|c|c|c|c|c|c|c|c|c|c|c|c|c|c|}\cline{5-17}
\multicolumn{4}{c|}{\centering \scriptsize{}} &
\multicolumn{13}{c|}{Individual Color Representations}\\\cline{5-17}
\multicolumn{4}{p{0.3cm}|}{\centering {}}
&$R$&$G$&$B$&$nr$&$ng$&$O_1$&$O_2$&$L$&$a$&$b$&$H$&$S$&$V$\\\hline
\multirow{5}{*}{\centering\begin{sideways}\centering{Non-param.}\end{sideways}}&\multicolumn{3}{l|}{{Nearest neighbor}}&
$74.2$&$72.7$&$87.8$&$84.4$&$56.6$&$72.3$&$88.3$&$70.6$&$63.8$&$90.1$&$78.9$&$87.1$&$83.5$\\
\cline{2-17}&\multirow{4}{*}{\centering\begin{sideways}\centering{\scriptsize{model-based}}\end{sideways}}&\multicolumn{1}{c|}{{$64$}}&{S}&
$73.7$&$74.1$&$74.9$&$76.4$&$73.5$&$75.4$&$75.4$&$73.8$&$76.6$&$77.9$&$79.7$&$76.4$&$73.7$\\
\cline{4-17}&&\multicolumn{1}{c|}{{bins}}&{S\&N}&
$76.1$&$76.7$&$78.2$&$76.4$&$73.5$&$75.3$&$75.5$&$77.4$&$76.7$&$77.9$&$79.6$&$76.8$&$77.3$\\
\cline{3-17}&&{{$100$}}&{S}&
$73.4$&$73.8$&$74.5$&$77.0$&$74.8$&$76.3$&$76.6$&$73.5$&$77.0$&$78.5$&$79.7$&$76.9$&$73.4$\\
\cline{4-17}&&{bins}&{S\&N}&
$76.1$&$76.6$&$78.2$&$77.0$&$74.7$&$76.3$&$76.7$&$77.4$&$77.1$&$78.5$&$79.7$&$77.4$&$77.3$\\
\hline\multirow{13}{*}{\centering\begin{sideways}\centering{Parametric}\end{sideways}}&\multicolumn{3}{l|}{{Gaussian}}&
$75.7$&$83.1$&$87.8$&$87.9$&$56.3$&$67.6$&$92.0$&$78.8$&$68.4$&$92.5$&$85.5$&$92.8$&$83.6$\\
\cline{2-17}&\multicolumn{3}{l|}{{Robust Gaussian}}&
$75.7$&$83.1$&$87.8$&$87.9$&$56.5$&$67.6$&$92.0$&$78.8$&$68.5$&$92.5$&$85.5$&$92.8$&$83.6$\\
\cline{2-17}&\multicolumn{2}{l|}{\multirow{3}{*}{MoG}}&{2}&
$74.2$&$74.4$&$76.2$&$75.5$&$75.6$&$75.8$&$77.4$&$74.7$&$76.5$&$78.7$&$78.6$&$76.0$&$74.8$\\
\cline{4-17}&\multicolumn{2}{l|}{}&{4}&
$79.3$&$80.1$&$81.8$&$80.0$&$78.2$&$78.3$&$79.3$&$80.2$&$79.9$&$78.7$&$81.3$&$78.9$&$80.5$\\
\cline{4-17}&\multicolumn{2}{l|}{}&{opt.}&
$75.8$&$83.1$&$87.5$&$88.3$&$56.3$&$67.6$&$92.0$&$78.8$&$67.9$&$92.2$&$86.2$&$92.8$&$83.6$\\
\cline{2-17}&\multicolumn{3}{l|}{{SVD}}&
$75.8$&$83.2$&$87.7$&$88.2$&$55.7$&$68.4$&$91.9$&$78.8$&$67.5$&$92.5$&$83.9$&$92.9$&$83.7$\\
\cline{2-17}&\multicolumn{3}{l|}{{PCA}}&
$--$&$--$&$--$&$--$&$--$&$--$&$--$&$--$&$--$&$--$&$--$&$--$&$--$\\
\cline{2-17}&\multicolumn{3}{l|}{{MPM}}&
$75.8$&$83.1$&$87.7$&$88.0$&$56.3$&$67.8$&$92.0$&$78.8$&$68.2$&$92.5$&$86.7$&$92.9$&$83.7$\\
\cline{2-17}&\multicolumn{3}{l|}{{MST}}&
$75.5$&$83.1$&$87.9$&$87.2$&$56.5$&$68.1$&$91.5$&$78.8$&$67.0$&$92.2$&$86.1$&$92.8$&$83.6$\\
\cline{2-17}&\multicolumn{3}{l|}{{dLP}}&
$75.8$&$83.2$&$87.7$&$88.2$&$55.7$&$68.4$&$91.8$&$78.8$&$67.5$&$92.5$&$83.3$&$92.9$&$83.6$\\
\cline{2-17}&\multicolumn{3}{l|}{{kmeans}}&
$75.4$&$83.1$&$87.6$&$87.3$&$56.3$&$68.3$&$91.4$&$78.7$&$68.7$&$92.3$&$84.8$&$92.8$&$83.7$\\
\cline{2-17}&\multicolumn{3}{l|}{{kcenter}}&
$75.3$&$83.0$&$88.0$&$87.5$&$56.8$&$68.1$&$91.3$&$78.8$&$66.9$&$92.1$&$84.8$&$92.8$&$83.6$\\
\cline{4-17}\hline
\end{tabular}
}
\\(a)\\
\end{center}

\begin{center}
{\small
\begin{tabular}{|p{0.1cm}|p{0.4cm}|p{0.3cm}|c|c|c|c|c|c|c|c|c|c|}\cline{5-10}
\multicolumn{4}{c|}{} &\multicolumn{6}{c|}{Combination of Color Representations}\\\cline{5-10}
\multicolumn{4}{p{0.3cm}|}{\centering {}}
&{$RGB$}&$nrng$&$O_1O_2$&$Lab$&$HSV$&$HS$\\\hline
\multirow{5}{*}{\centering\begin{sideways}\centering{Non-param.}\end{sideways}}&\multicolumn{3}{l|}{{Nearest neighbor}}&
$83.1$&$88.2$&$91.1$&$86.6$&$85.1$&$91.6$\\
\cline{2-10}&\multirow{4}{*}{\centering\begin{sideways}\centering{\scriptsize{model-based}}\end{sideways}}&\multicolumn{1}{c|}{{$64$}}&{S}&
$58.0$&$54.2$&$68.2$&$54.9$&$53.8$&$53.2$\\
\cline{4-10}&&\multicolumn{1}{c|}{{bins}}&{S\&N}&
$70.6$&$55.4$&$68.3$&$53.4$&$62.8$&$63.0$\\
\cline{3-10}&&{{$100$}}&{S}&
$55.8$&$54.0$&$68.9$&$53.9$&$53.2$&$53.6$\\
\cline{4-10}&&{bins}&{S\&N}&
$61.4$&$55.7$&$59.5$&$53.7$&$57.6$&$62.3$\\
\hline\multirow{13}{*}{\centering\begin{sideways}\centering{Parametric}\end{sideways}}&\multicolumn{3}{l|}{{Gaussian}}&
$90.9$&$90.0$&$92.9$&$90.9$&$91.6$&$93.2$\\
\cline{2-10}&\multicolumn{3}{l|}{{Robust Gaussian}}&
$90.0$&$89.4$&$93.0$&$91.4$&$79.1$&$\mathbf{93.4}$\\
\cline{2-10}&\multicolumn{2}{l|}{\multirow{3}{*}{MoG}}&{2}&
$85.6$&$85.8$&$83.3$&$86.1$&$86.3$&$84.8$\\
\cline{4-10}&\multicolumn{2}{l|}{}&{4}&
$88.3$&$88.6$&$84.8$&$88.3$&$88.5$&$86.3$\\
\cline{4-10}&\multicolumn{2}{l|}{}&{Opt.}&
$84.5$&$90.8$&$85.7$&$75.4$&$84.4$&$88.2$\\

\cline{2-10}&\multicolumn{3}{l|}{{SVD}}&
$81.7$&$91.0$&$91.9$&$85.3$&$84.6$&$92.7$\\
\cline{2-10}&\multicolumn{3}{l|}{{PCA}}&
$91.6$&$--$&$--$&$91.2$&$89.3$&$--$\\
\cline{2-10}&\multicolumn{3}{l|}{{MPM}}&
$81.7$&$90.9$&$92.0$&$85.3$&$84.6$&$92.8$\\
\cline{2-10}&\multicolumn{3}{l|}{{MST}}&
$82.8$&$90.9$&$92.1$&$86.5$&$85.0$&$93.0$\\
\cline{2-10}&\multicolumn{3}{l|}{{dLP}}&
$81.7$&$91.1$&$91.9$&$85.3$&$84.6$&$92.7$\\
\cline{2-10}&\multicolumn{3}{l|}{{kmeans}}&
$82.7$&$91.1$&$92.1$&$86.2$&$85.1$&$93.0$\\
\cline{2-10}&\multicolumn{3}{l|}{{kcenters}}&
$82.6$&$91.5$&$92.2$&$86.3$&$84.9$&$93.1$\\
\cline{3-10}\hline
\end{tabular}
}
\\(b)
\end{center}
\end{table*}

\subsection{Results}
\label{subsect:evals}

\begin{figure*}[htpb!]
\begin{center}
\begin{tabular}{cc}
\multicolumn{2}{c}{\includegraphics[width=\textwidth]{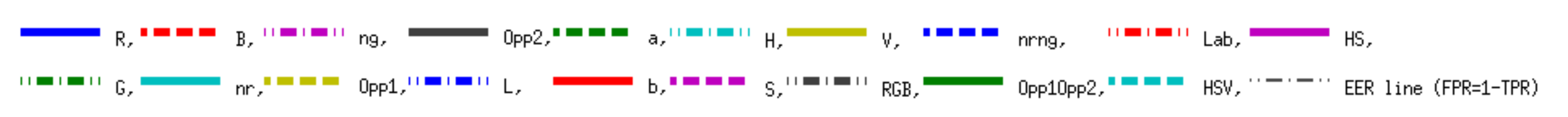}}\\
\hspace{-0.3cm}\includegraphics[width=0.495\textwidth,height=4.5cm]{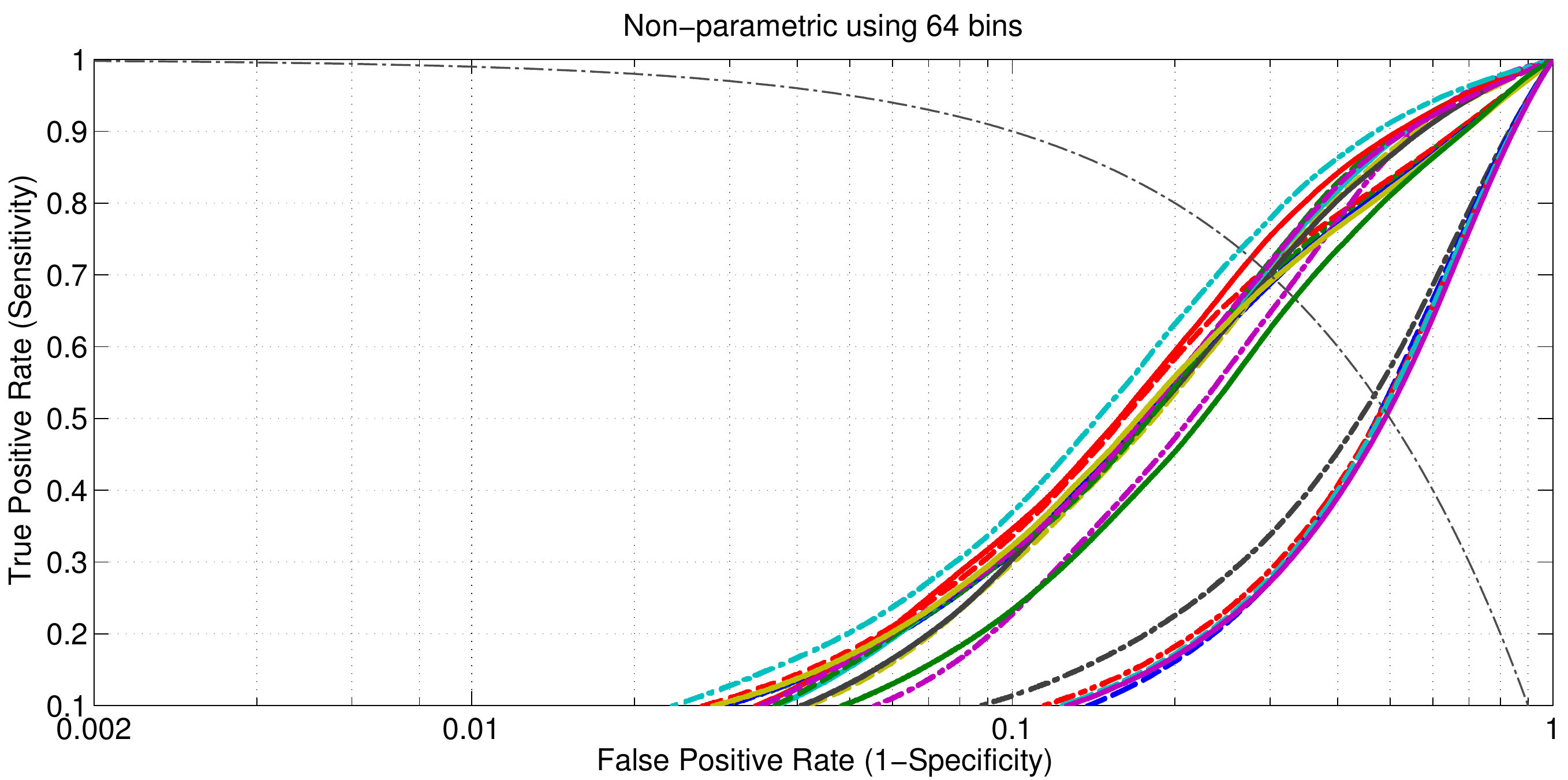}&
\hspace{-0.3cm}\includegraphics[width=0.495\textwidth,height=4.5cm]{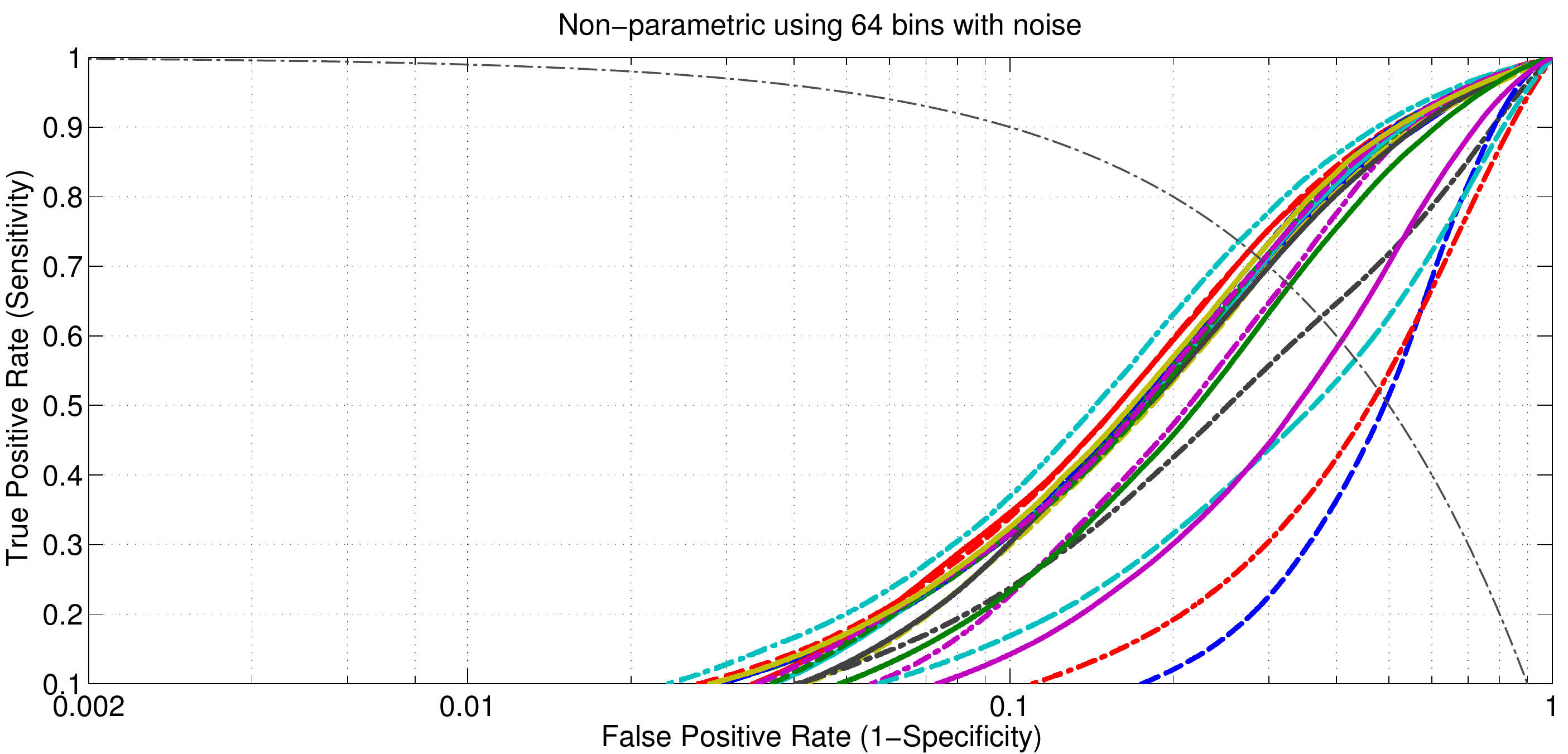}\\
\hspace{-0.3cm}\includegraphics[width=0.495\textwidth,height=4.5cm]{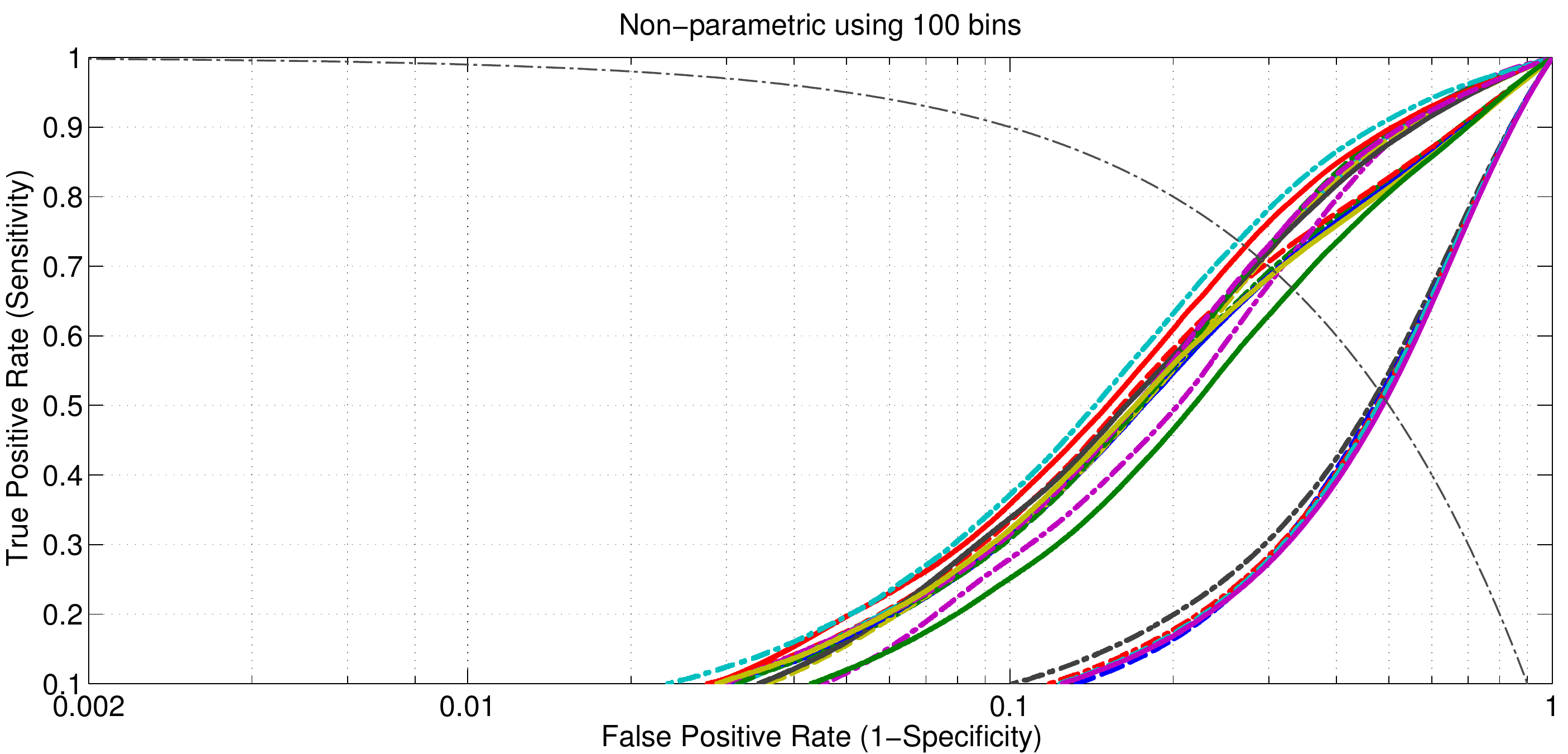}&
\hspace{-0.3cm}\includegraphics[width=0.495\textwidth,height=4.5cm]{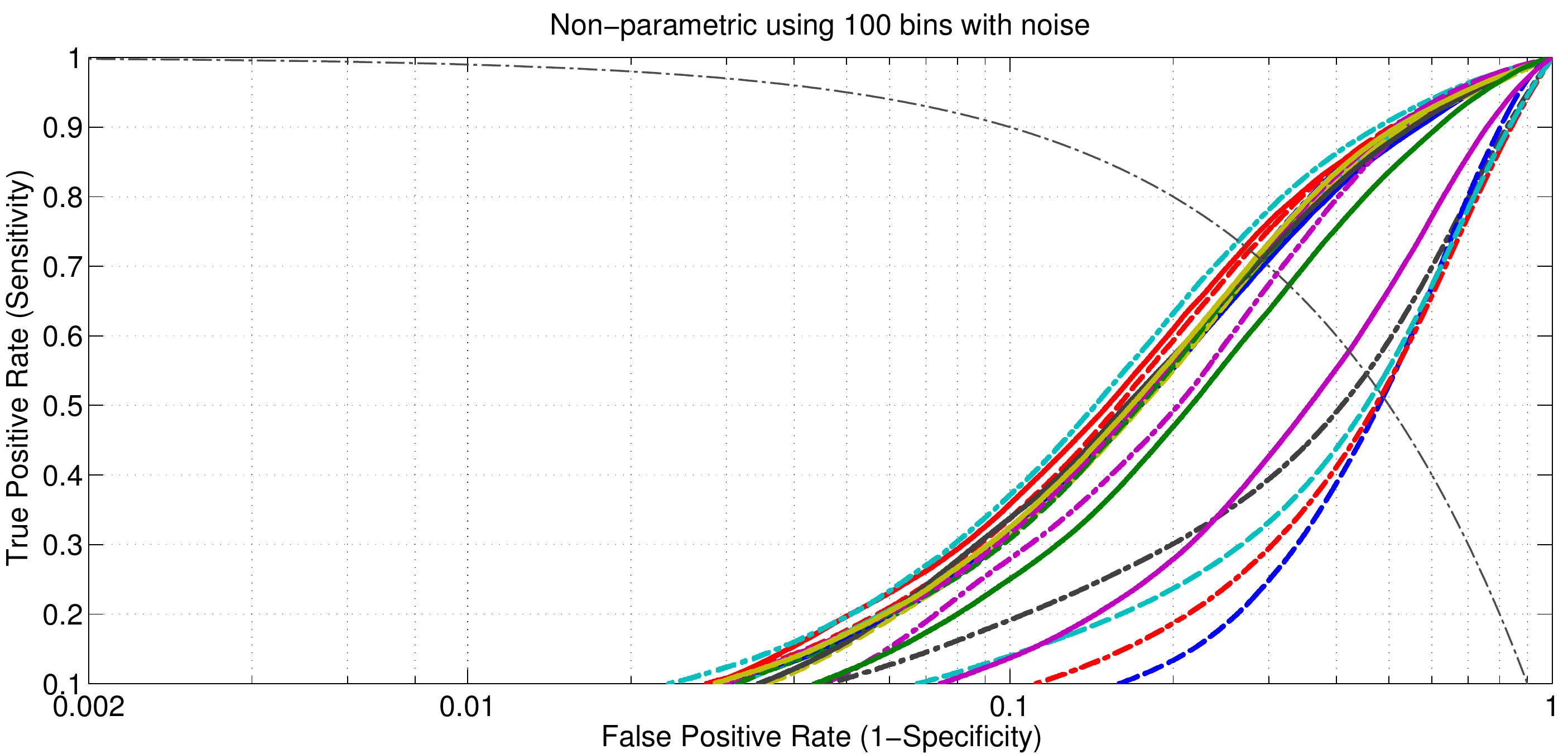}\\
\end{tabular}
\end{center}
 \vspace{-0.5cm}
\caption{ROC curves for the different instances of the model-based classifier. As shown, the classifier is stable to variations in the number of bins and, in general, extending the training set with noisy instances improve results (right column).}\vspace{-0.5cm}
\label{fig:ROCcurvesMODEL}
\end{figure*}

\begin{figure*}[htpb!]
\begin{center}
\begin{tabular}{cc}
\multicolumn{2}{c}{\includegraphics[width=\textwidth]{legend.pdf}}\\
\vspace{-0.4cm}
\hspace{-0.3cm}\includegraphics[width=0.495\textwidth,height=4.5cm]{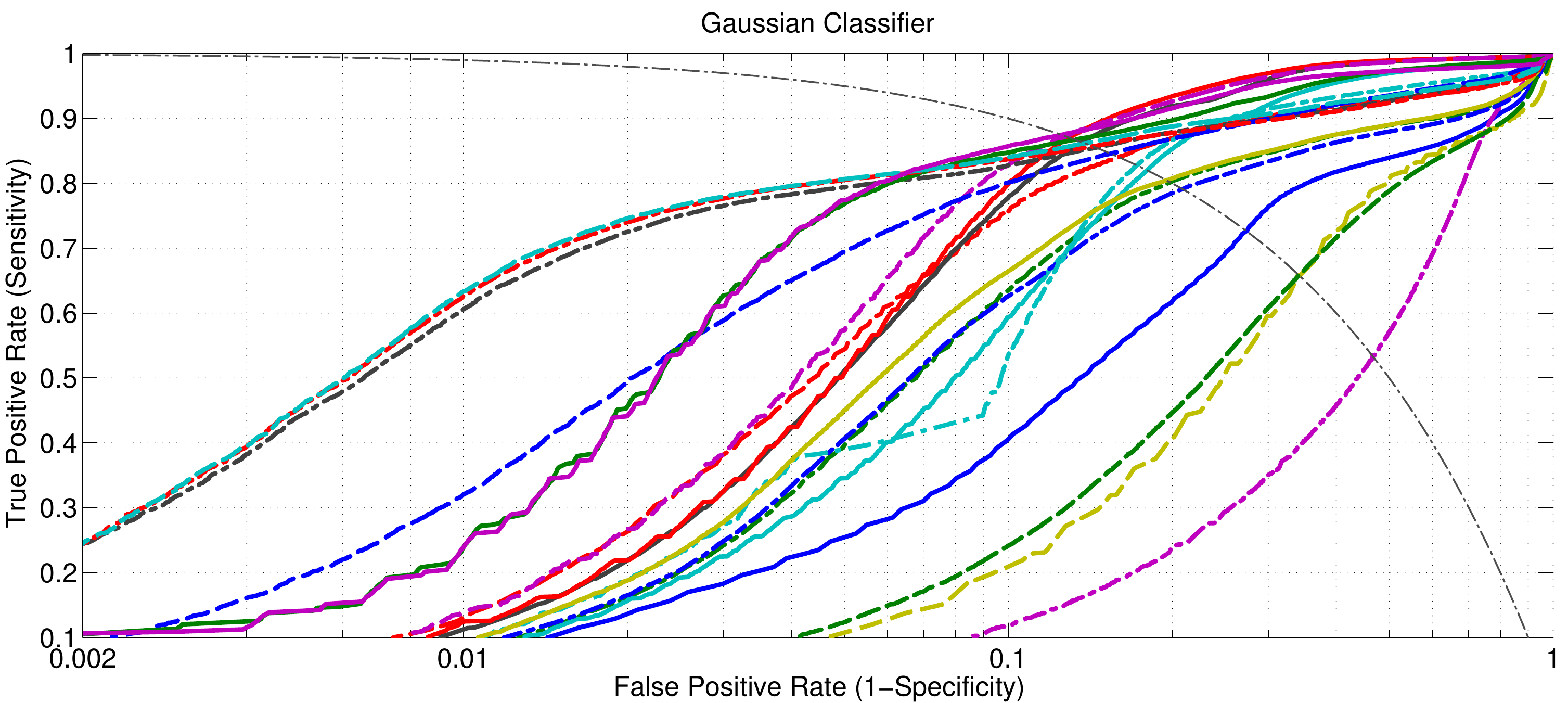}&
 \hspace{-0.3cm}\includegraphics[width=0.495\textwidth,height=4.5cm]{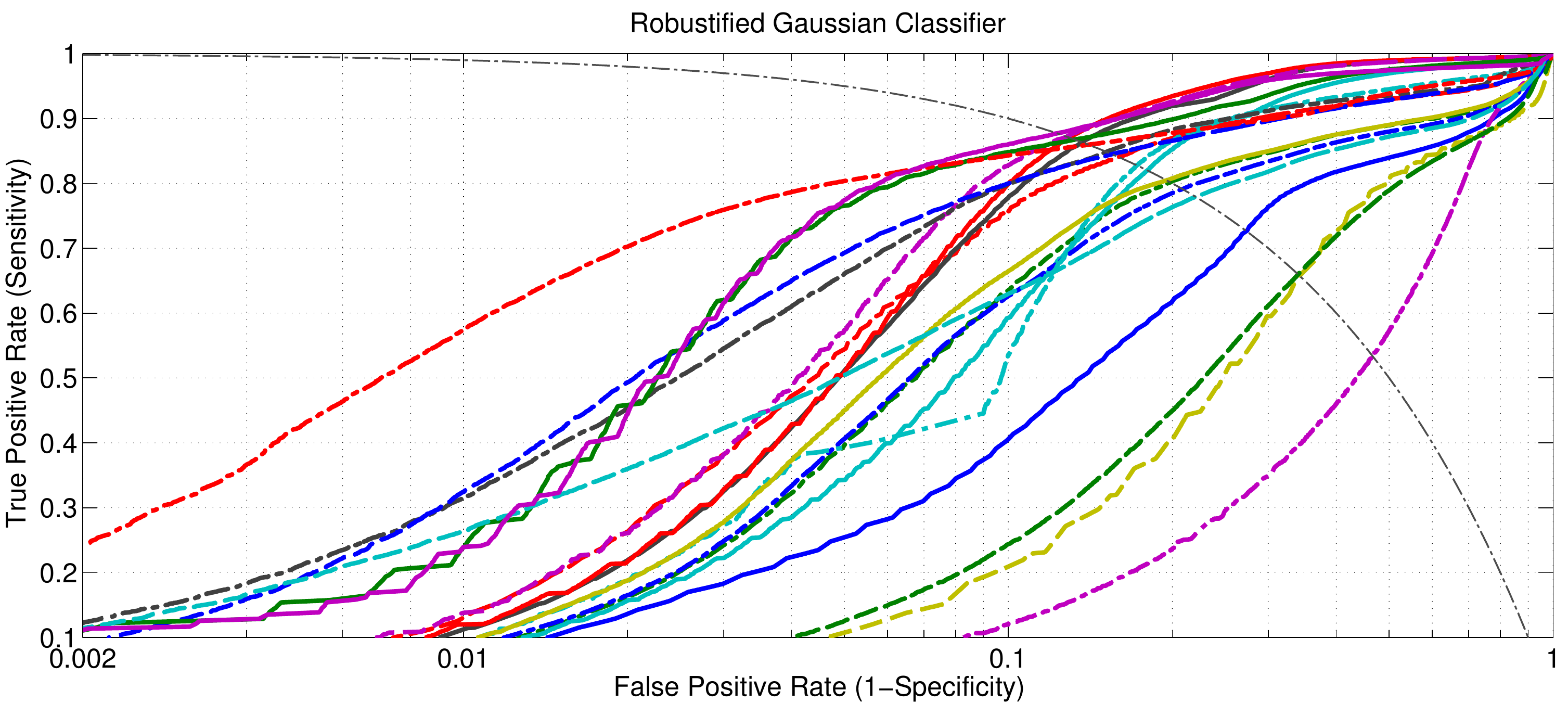}\\
 \hspace{-0.3cm}\includegraphics[width=0.495\textwidth,height=4.5cm]{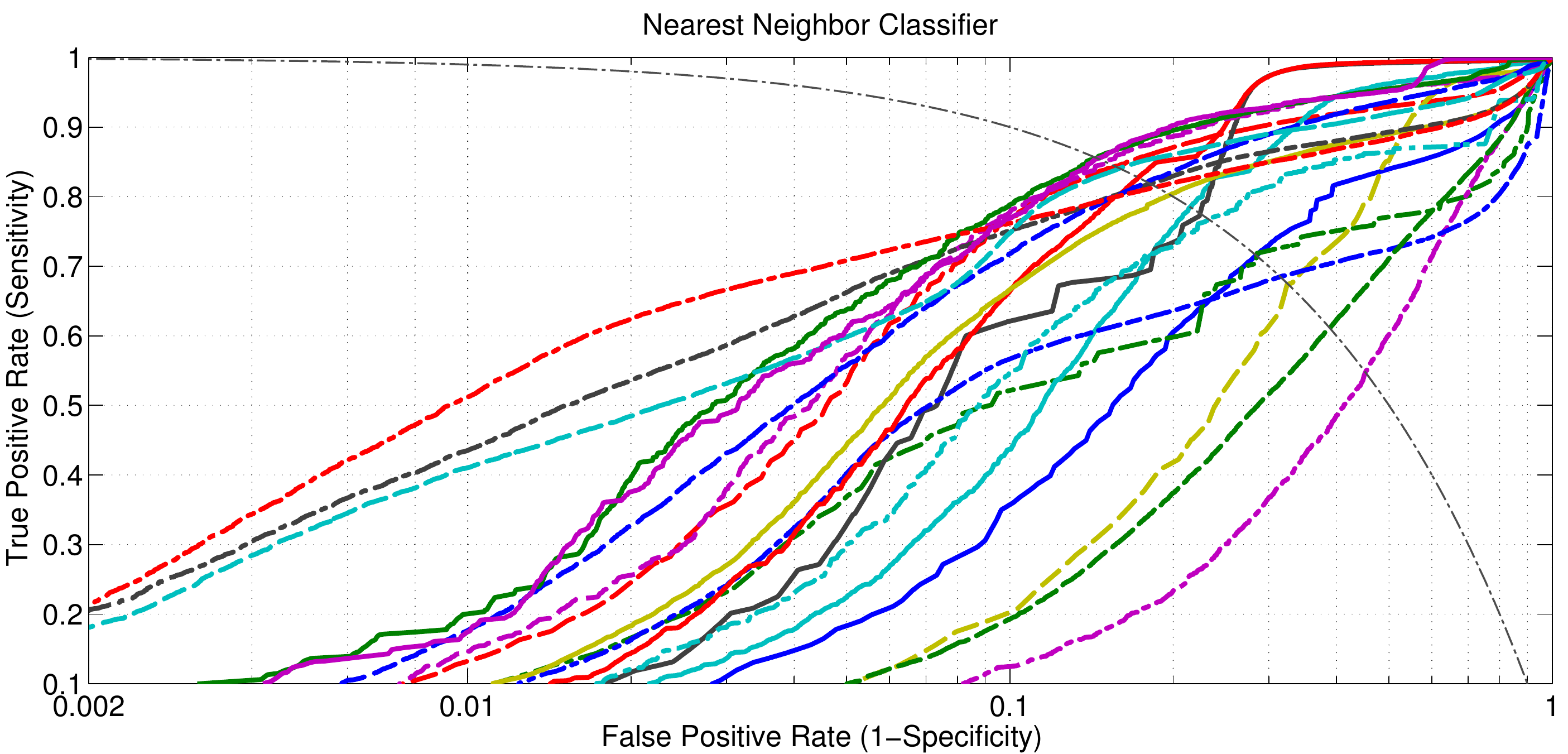}&
 \hspace{-0.3cm}\includegraphics[width=0.495\textwidth,height=4.5cm]{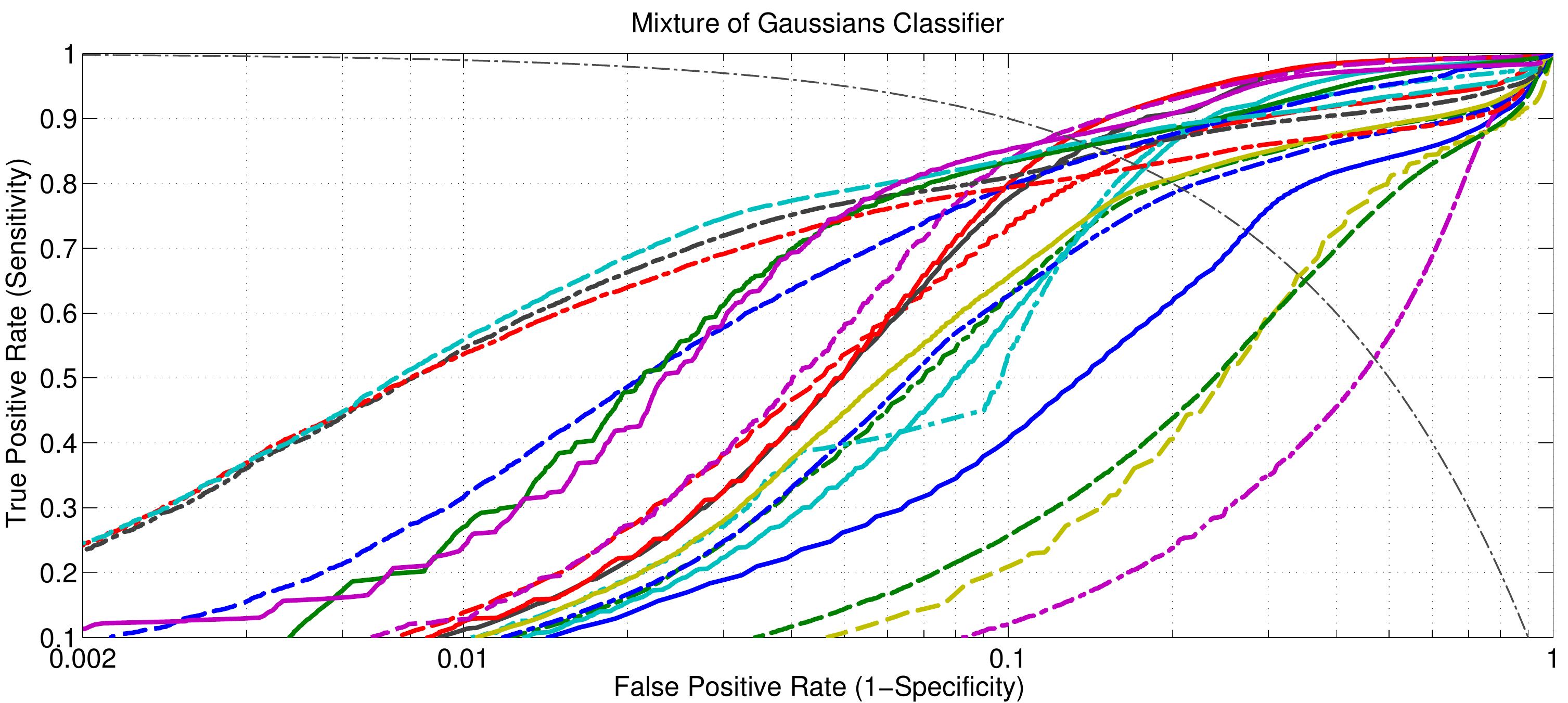}\\
 \hspace{-0.3cm}\includegraphics[width=0.495\textwidth,height=4.5cm]{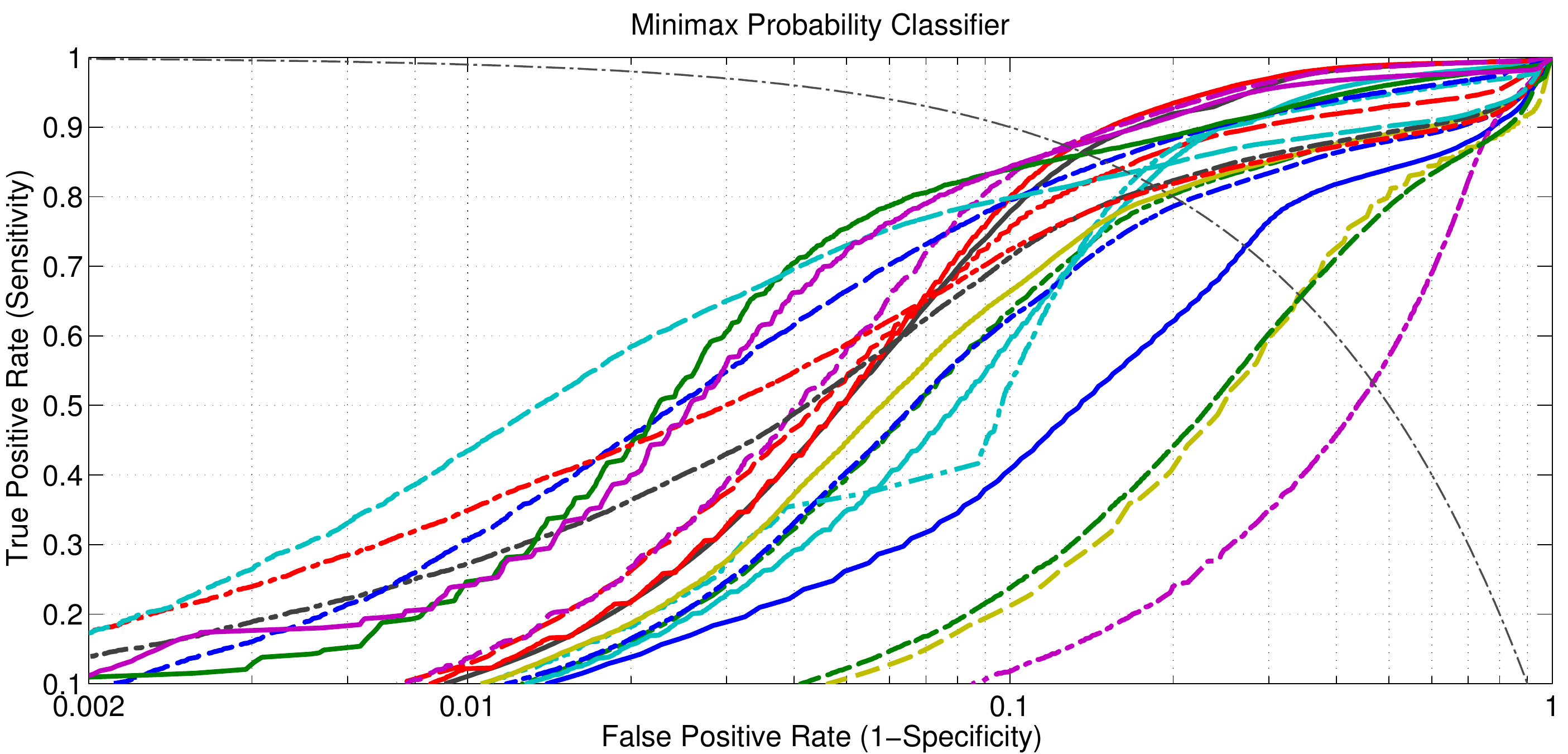}&
 \hspace{-0.3cm}\includegraphics[width=0.495\textwidth,height=4.5cm]
 {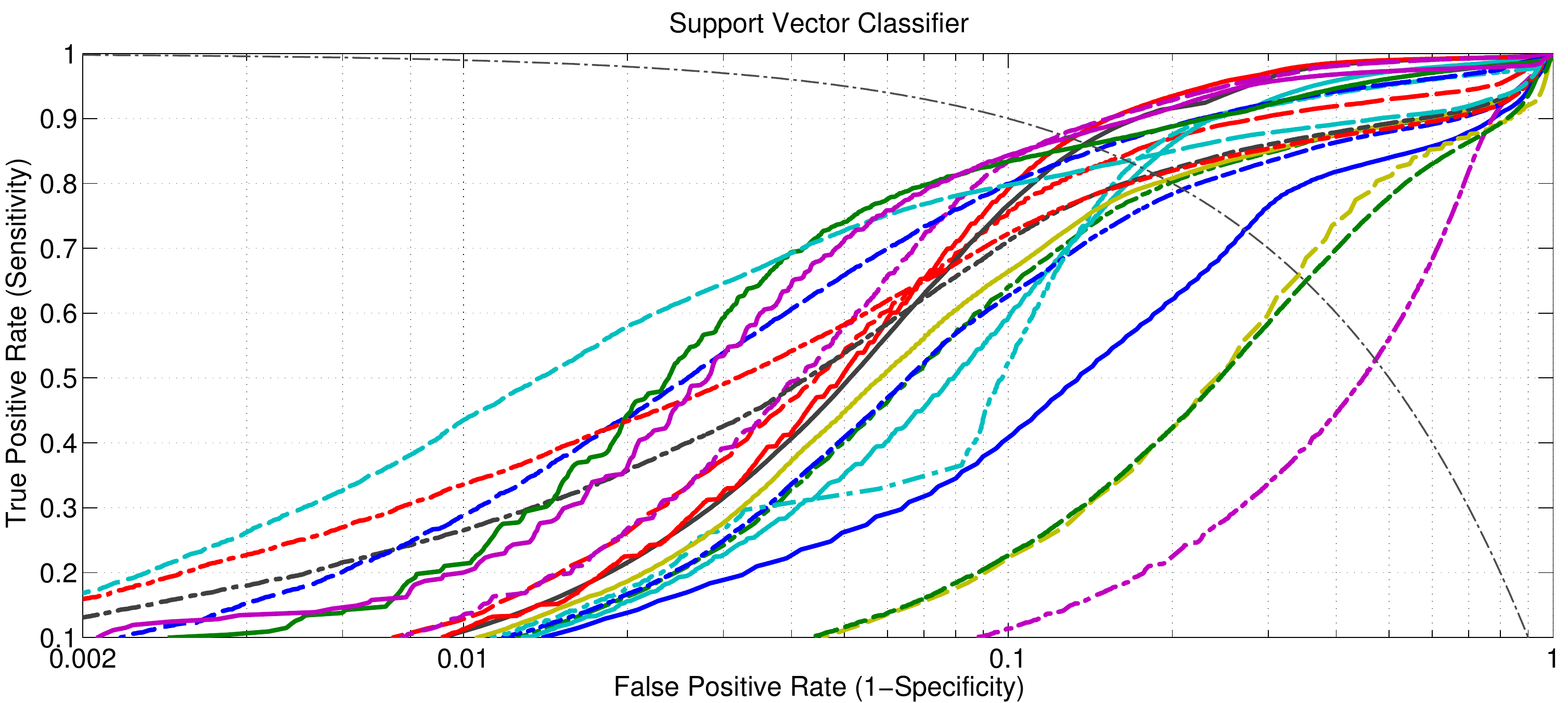}\\
 \hspace{-0.3cm}\includegraphics[width=0.495\textwidth,height=4.5cm]{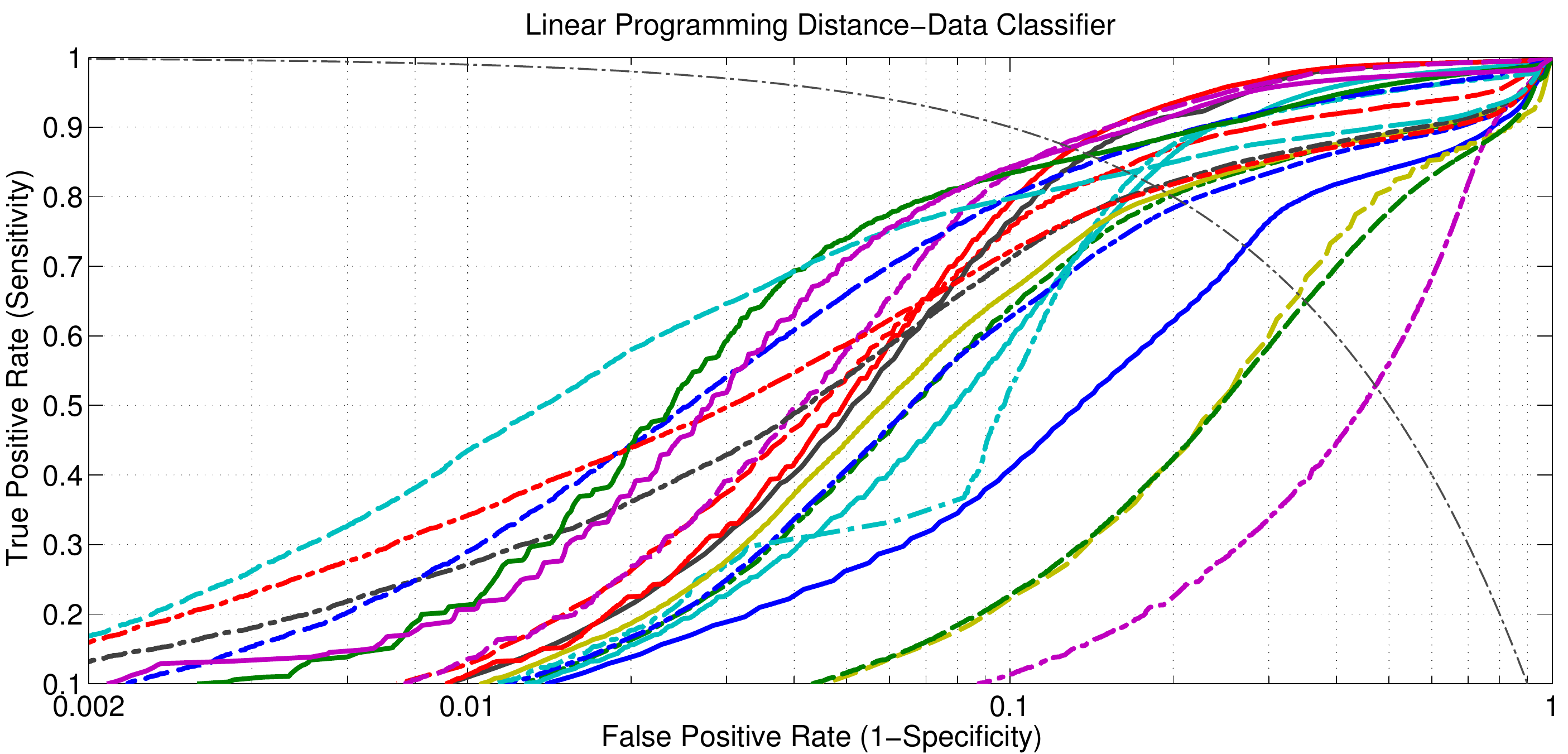}&
 \hspace{-0.3cm}\includegraphics[width=0.495\textwidth,height=4.5cm]{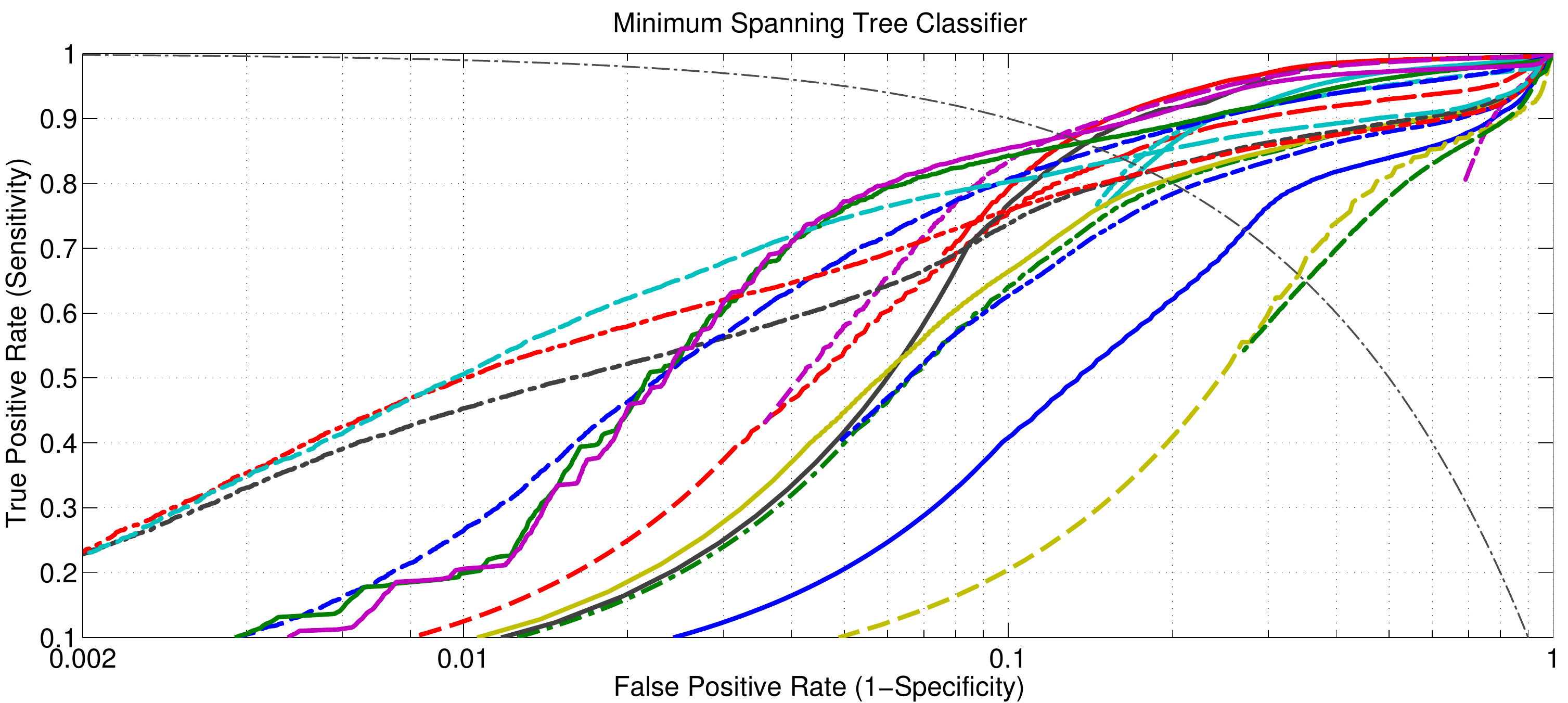}\\
 \hspace{-0.3cm}\includegraphics[width=0.495\textwidth,height=4.5cm]
 {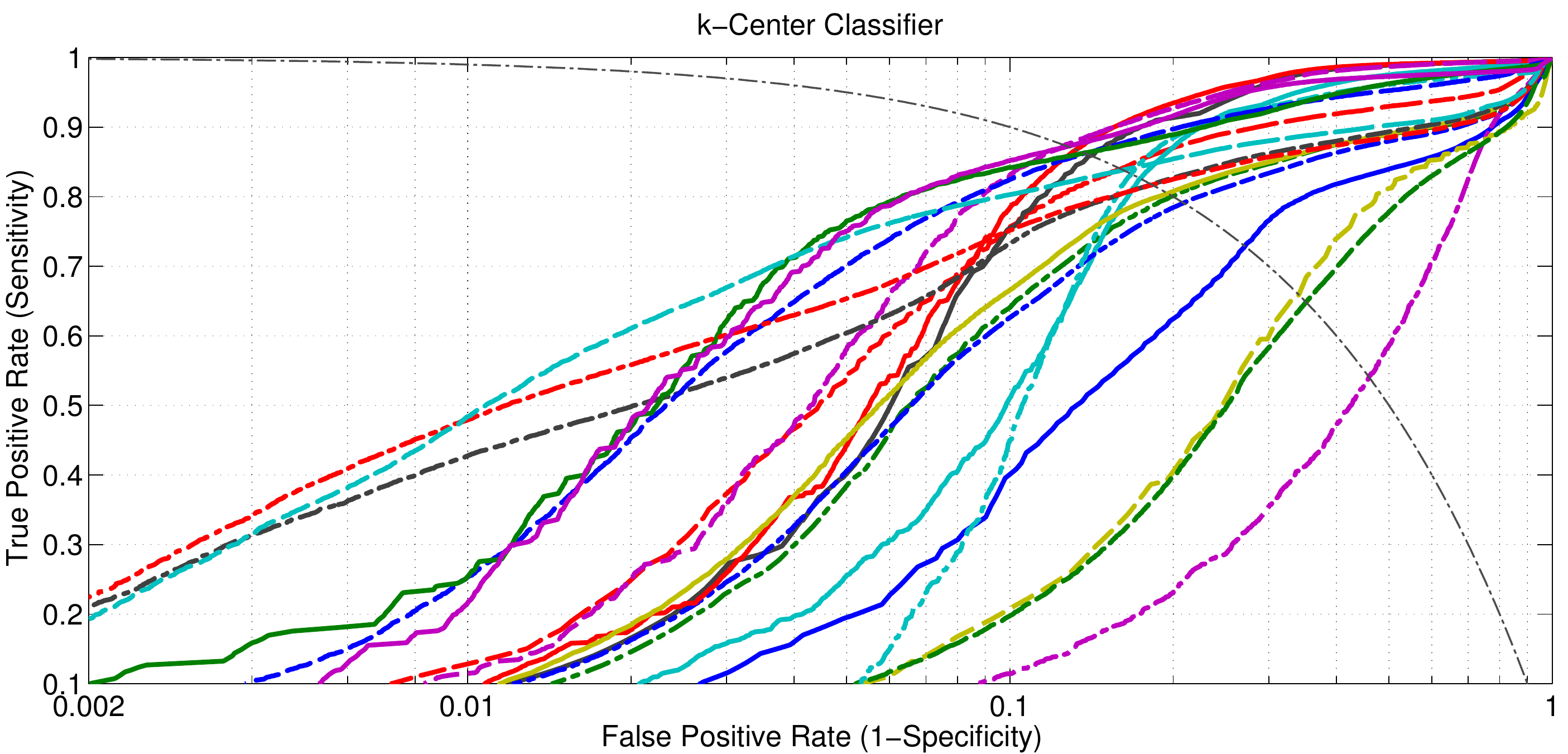}&
 \hspace{-0.3cm}\includegraphics[width=0.495\textwidth,height=4.5cm]
 {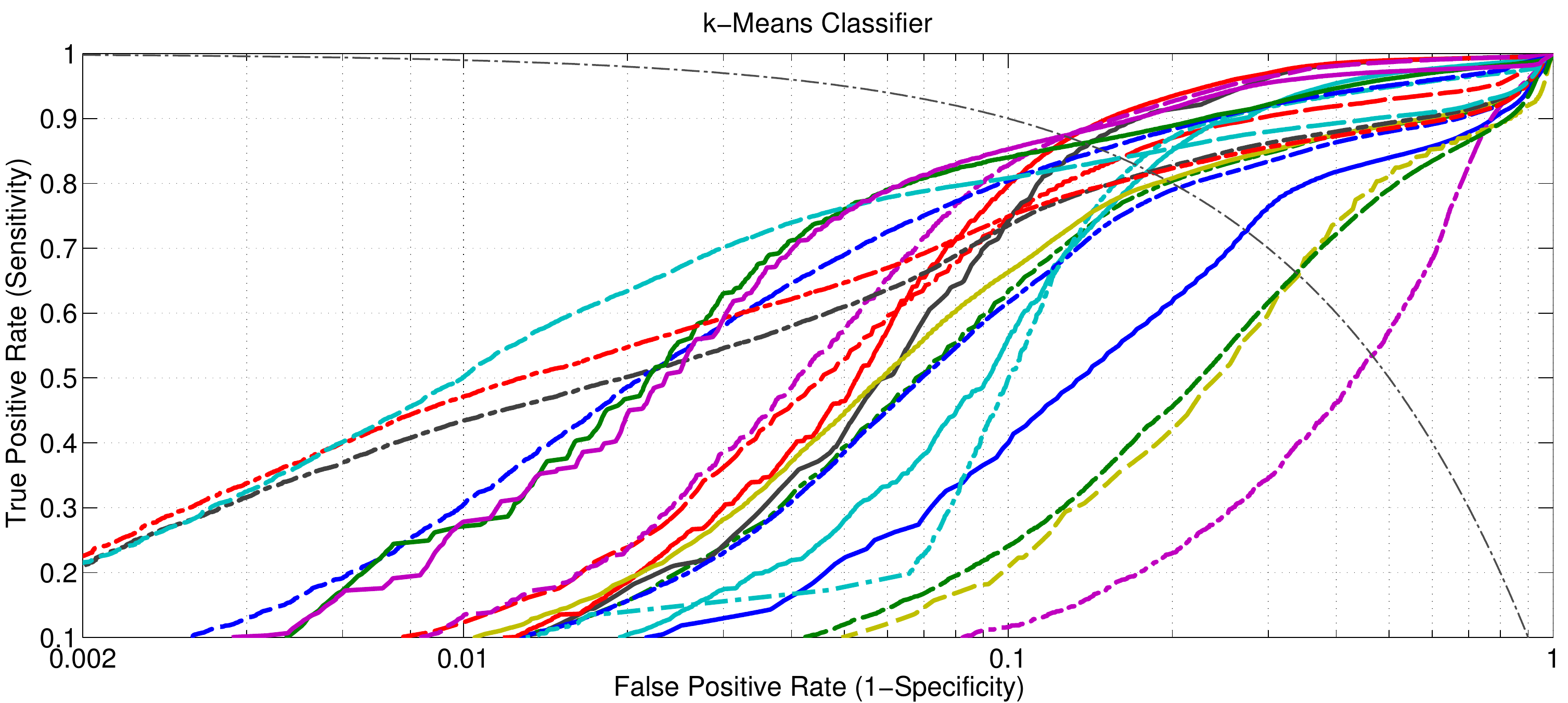}\\
 \end{tabular}
 \end{center}
 \vspace{-0.4cm}
 \caption{ROC curves for different instances of single-class classifiers using single color planes and their most common combinations. Quatitative evaluations are summarized in~\tab{tab:evalNaming}.}
 \label{fig:ROCcurves} \vspace{-0.5cm}
 \end{figure*}

The summary of the AUC values resulting from combining the $19$
different color representations and $17$ instances of single class
classifiers is listed in Table~\ref{tab:evalNaming}. ROC curves
for the different instances of the model-based classifier are
shown in~\fig{fig:ROCcurvesMODEL} and representative ROC curves
for the rest of classifiers are shown in~\fig{fig:ROCcurves}.
From~\fig{fig:ROCcurvesMODEL}, we can derive that the stability of
the model-based classifier with respect to the number of bins used
to build the histogram. The relative low performance of this
model-based classifier tends to improve by extending the training
set when using noisy samples. This is probably due to the lack of
training samples representing the target (road) class. Therefore,
adding noisy samples improves the variety of the training data.
Note the performance drop of this model-based classifier when
considering multiple color planes. This suggests that the joint
distribution of these color planes can not capture the road
appearance using only a few training samples. These results could
be improved by considering the likelihood provided by each color
plane independently.

As expected, the PCA classifier can not perform with single color
planes. PCA is based on data covariance matrix. Therefore, is not
suitable for single dimension input. Nevertheless, this classifier
provides outstanding performance when using three dimensional
input data. Further, besides model-based classifier on joint
distributions, the worst performance corresponds to the linear
programming classifier (dLP) in the $ng$ color space. As shown,
the performance of this color space is generaly low. This is
mainly due to an excess of invariance leading to higher false
positive rates (\ie, the model has not discriminative properties)
as shown in~\fig{fig:ROCcurves}. The use of $O_1$ as input to most
of the classifiers also provides low performance. This opponent
component is a combination or $R$ and $G$ color planes. In
contrast, $O_2$ and $b$ color planes provides good performance.
Worth noticing that these two color components are combinations of
$RGB$ color space and include certain amount of $B$. This suggests
that $B$ provides relevant information to balance invariance and
discriminative power. As shown in Table~\ref{tab:evalNaming},
among $RGB$ color planes, $B$ is the one providing higher
performance.

Interestingly, high performance is achieved when the input data contains $H$, and in particular, when the input data is $HS$. The performance drops when luminance is included (as $HSV$). This is mainly due to the lack of inviariance in the luminance color plane. Therefore, we can conclude that, for challenging situations, the use of luminance color space decreases the performance of the algorithms. Nevertheless, in real world driving situations these challenging conditions may only represent a small portion of the dataset (\eg, the dataset presented here is a subset of a large dataset with less-challenging images) and a different color plane may be more suitable. For instance, when single color planes are evaluated in general sequences $RGB$ outperforms the other color planes.

\section{Conclusions}
\label{sect:conclusions}
In this paper, we introduced a comprehensive evaluation combining $19$ color representations with $17$ different single class-classifiers for road detection. Experiments were conducted on a new set of road images comprising $755$ manually annotated images. From the results, we conclude that combining multiple color representations using a parametric classifier outperforms the accuracy of single color representations. Moreover, in this dataset, learning a robustified Gaussian model in a color space using both saturation and hue yields highest accuracy.

{\small
\bibliographystyle{IEEEbib}
\bibliography{JM_GEN_PAMI_short2}\vspace{-0.75cm}
}
\begin{IEEEbiography}	[{\includegraphics[width=1in,height=1.25in,clip,keepaspectratio]{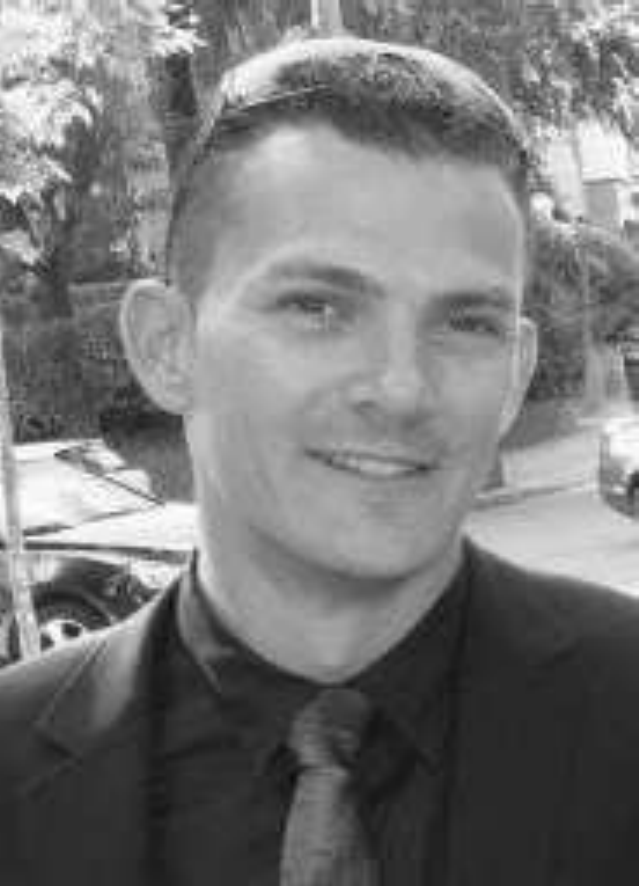}}]{Jose~M.~\'Alvarez}
is currently a senior researcher at NICTA and a research fellow at the Australian National University. Previously, he was a visiting postdoctoral researcher at the Computational and Biological Learning Group at New York University with Prof. Yann LeCun. During his Ph.D. he was a visiting researcher at the University of Amsterdam and Volkswagen research. His main research interests include large scale scene understanding, feature learning, road detection, color, photometric invariance. He is associate editor for the IEEE Trans. on Intel. Transportation Systems and member of the IEEE.
\vspace{-0.5cm}
\end{IEEEbiography}

\begin{IEEEbiography}[{\includegraphics[width=1in,height=1.25in,clip,keepaspectratio]{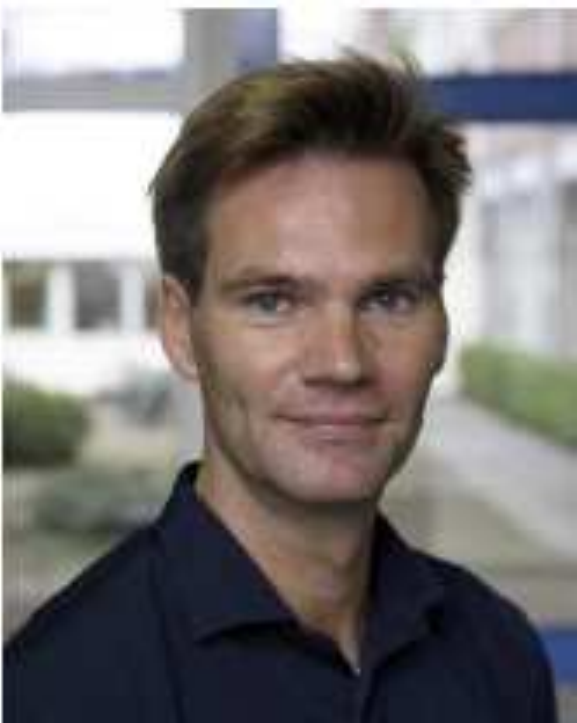}}]{Theo Gevers} is an Associate Professor of Computer Science at the University of Amsterdam, The Netherlands, where he is also teaching director of the M.Sc. of Artificial Intelligence. He currently holds a
VICI-award (for excellent researchers) from the Dutch Organisation
for Scientific Research. His main research interests are in the
fundamentals of content-based image retrieval, color image
processing and computer vision specifically in the theoretical
foundation of geometric and photometric invariants. He is chair of
various conferences and he is associate editor for the IEEE Trans.
on Image Processing. He is member of the IEEE.\vspace{-0.5cm}
\end{IEEEbiography}
\begin{IEEEbiography}[{\includegraphics[width=1in,height=1.25in,clip,keepaspectratio]{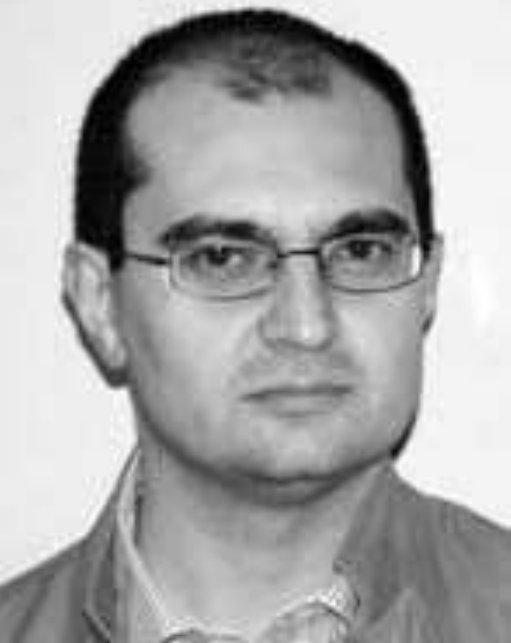}}]{Antonio~M.~L\'opez}
received his PhD degree in 2000 from the Universitat
Aut\`onoma de Barcelona (UAB) where, currently, he is an associate
professor. In 1996, he participated in the foundation of the
Computer Vision Center at the UAB, where he has held different
institutional responsibilities, presently being the responsible
for the research group on advanced driver assistance systems by
computer vision. He has been responsible for public and private
projects, and is a coauthor of more than $100$ papers in the
field of computer vision. He is a member of the IEEE.
\end{IEEEbiography}

\end{document}